\documentclass[a4paper,11pt,oneside]{article}

\usepackage{amsmath,amsthm,amsfonts,amssymb,amscd}
\usepackage{mathtools}
\usepackage{float}
\usepackage{microtype}

\usepackage{setspace}
\usepackage{subfig} 
\usepackage{enumerate}
\usepackage{booktabs}
\usepackage[noend]{algpseudocode}
\usepackage[normalem]{ulem}
\usepackage{graphicx,bm,xcolor}
\usepackage{algorithm}
\usepackage{algpseudocode}
\usepackage{hyperref} 
\usepackage{caption}
\usepackage[T1]{fontenc}
\usepackage{t1enc}
\usepackage[polish,german,english]{babel}
\usepackage{verbatim}
\usepackage{diagbox}
\usepackage{colortbl}
\usepackage{siunitx}
\usepackage{mathdots} 
\usepackage{bbm}
\usepackage{cancel} 
\usepackage{pgf,tikz,pgfplots}
\usepackage{mathrsfs}
\usetikzlibrary{arrows}
\usetikzlibrary{shapes.misc}
\usetikzlibrary{shapes.geometric, arrows,chains}
\usetikzlibrary{positioning}
\usetikzlibrary{decorations.pathreplacing} 
\usetikzlibrary{calc}
\usepackage{tikz-cd} 

\usepackage{xcolor} 

\usepackage{url}
\usepackage{authblk}

\algnewcommand{\algorithmicgoto}{\textbf{go to}}
\algnewcommand{\Goto}[1]{\algorithmicgoto~\ref{#1}}

\usepackage[colorinlistoftodos,prependcaption,textsize=scriptsize,color=red!40!white]{todonotes}\usepackage{totcount}

\definecolor{orange}{RGB}{230, 159, 0}
\definecolor{skyblue}{RGB}{86, 180, 233}
\definecolor{yellow}{RGB}{240, 228, 66}
\definecolor{blue}{RGB}{0, 114, 178}
\definecolor{vermillion}{RGB}{213, 94, 0}

\usepackage[top=2cm, bottom=2cm, left=2cm, right=2cm]{geometry}

\theoremstyle{plain} 
\newtheorem{theorem}{Theorem}[section]

\newtheorem{form}[theorem]{Formulation}

\theoremstyle{definition} %

\theoremstyle{remark} %

\definecolor{brewer1}{HTML}{A6CEE3}
\definecolor{brewer2}{HTML}{1F78B4}
\definecolor{brewer3}{HTML}{B2DF8A}
\definecolor{brewer4}{HTML}{33A02C}

\def\rightfanD{\mathrel{\hbox{\rlap{$-$}$<$}}}
\let\rightfanT\rightfanD
\def\rightfanS{\mathrel{\hbox{\rlap{$\scriptstyle-$}$\scriptstyle<$}}}
\def\rightfanSS{%
\mathrel{\hbox{\rlap{$\scriptscriptstyle-$}$\scriptscriptstyle<$}}}

\let\crowr\rightfan     

\begin{document}


\title{
  Generalization capabilities of MeshGraphNets \\ to unseen geometries for fluid dynamics
}

\author[1]{Robin Schmöcker}
\author[2]{Alexander Henkes}
\author[1]{Julian Roth}
\author[1,3]{Thomas Wick}

\affil[1]{Leibniz Universit\"at Hannover, Institut f\"ur Angewandte
  Mathematik, AG Wissenschaftliches Rechnen, Welfengarten 1, 30167 Hannover, Germany}
\affil[2]{Computational Mechanics Group, ETH Zürich, Switzerland}
\affil[3]{Universit\'e Paris-Saclay, CentraleSup\'elec, ENS Paris-Saclay, CNRS, LMPS - Laboratoire de Mécanique Paris-Saclay,
91190 Gif-sur-Yvette, France}

\date{}
\maketitle


\begin{abstract}
  This works investigates the generalization capabilities of MeshGraphNets (MGN) [Pfaff et al. Learning Mesh-Based Simulation
  with Graph Networks. ICML 2021] to unseen geometries for fluid dynamics, e.g. predicting the flow around a new obstacle that was not part of the training data. For this purpose, we create a new benchmark dataset for data-driven computational fluid dynamics (CFD) which extends DeepMind's flow around a cylinder dataset by including different shapes and multiple objects. We then use this new dataset to extend the generalization experiments conducted by DeepMind on MGNs by testing how well an MGN can generalize to different shapes. In our numerical tests, we show that MGNs can sometimes generalize well to various shapes by training on a dataset of one obstacle shape and testing on a dataset of another obstacle shape. 
\end{abstract}


\section{Introduction}
\label{sec_intro}

For many engineering applications, we need to have fast and accurate simulations of fluid dynamics.
In this work we focus on the modeling of fluids by the incompressible Navier-Stokes equations~\cite{galdi2011, temam2001, Rannacher2000, GiRa1986, turek1999, Glowinski2003, ElSiWa14, John2016}.
With regard to numerical modeling, various spatial discretization 
techniques are known such as the finite volume method~\cite{MouMaDa16} 
and the finite element method~\cite{GiRa1986}. Temporal discretization 
is often based on finite difference schemes~\cite{Glowinski2003,turek1999} or 
space-time discretizations~\cite{MiRaHaTe91,BeRa12,Anselmann2023}.
The arising nonlinear and linear systems
are often solved with iterative techniques or multigrid schemes.
Implementations on CPUs can be computationally expensive and can take a long time to run.
To overcome this, we can use parallelization to multiple processors or GPUs~\cite{TeBeMiJo92,John1999, ExaDG, Anselmann2023} or only refine the mesh in regions where the solution is not accurate enough \cite{BeRa2001, Braack2006, BeRa12, RoThiKoeWi2024}.
Nevertheless, even though this can speed up the simulation, fluid simulations can still be prohobitively expensive for multi-query problems, e.g. different material parameters, different geometries, or different boundary conditions.

Therefore, for real-time simulations or multi-query problems, we need to find a way to speed up the simulation even further, e.g. by means of data-driven methods.
A popular method to reduce the computational cost is to collect a large dataset of fluid simulations and project the dynamics onto a small set of eigen modes using the proper orthogonal decomposition \cite{Rozza2022CFD}.
On the one hand, using this kind of reduced order model is well-studied and we have error bounds for the reduced basis approximation \cite{Rozza2007Error}.
On the other hand, proper orthogonal decomposition is an intrusive method, which makes it difficult to implement for nonlinear problems \cite{Barrault2004AnI, Chaturantabut2010NonlinearMR, Chaturantabut2012ASS, Q-DEIM, Astrid2008MissingPE, Nguyen_patera_peraire_2008, Amsallem_Zahr_Farhat_2012} and changing meshes \cite{Graessle2019}.

Another approach is to use machine learning methods to predict the fluid dynamics \cite{kutz_2017}.
This kind of approaches are appealing since they are generally non-intrusive and easier to implement, but they can be difficult to interpret \cite{Erhan2009, Zeiler2014, Simonyan14a, bricken2023monosemanticity}, can be sensitive to the training data and can be difficult to generalize to unseen data \cite{novak2018sensitivity}.
Albeit all these limitations, neural networks are frequently used for regression in the latent space \cite{Ubbiali2018}, superresolution of coarse solutions \cite{Margenberg2022} or physics-informed approaches \cite{Cai2021}. 
More recently, convolutional neural networks have been used to learn the full solution field of fluid simulations \cite{Kochkov2021}. Despite their success in computer vision tasks \cite{AlexNet2012}, convolutional neural networks work best on cartesian grids and not for complex geometries, e.g. random obstacles in the flow field, where parts of the neural network input that is not part of the domain needs to be masked \cite{Grimm2023}.
To overcome this restriction of architectures that are based on cartesian grids, like convolutional neural networks or fourier neural operators, graph neural networks \cite{Bronstein2017, Bronstein2021Book, lengeling2021} can be used to directly work on the finite element mesh \cite{GINO2023}.

In this work, we will focus on a graph neural network-based solution introduced by Google DeepMind \cite{MGN_pre_paper, MGN}.
The method is based on the idea of encoding the system state into a mesh graph, applying a graph neural network to this graph to predict the change of the system, and then applying this change to the simulation state.
This method has found wide adoption for physics simulations with applications in weather forecasting (GraphCast) \cite{graphcast} and multiscale simulations for better efficiency \cite{MSMGN}.
The application of these neural networks is usually to inverse problems \cite{Allen2022} or for fast fluid simulations on unseen geometries \cite{Chen2021,wu2024Transolver}. The latter is the focus of this work.
We numerically investigate the generalization capabilities \cite{Tang2023TowardsUG} of MeshGraphNets to unseen geometries for fluid dynamics by predicting the flow around a new obstacle that was not part of the training data.
For this purpose we create new benchmark datasets \cite{bonnet2022} for the instationary incompressible Navier-Stokes equations that extend DeepMind's flow around a cylinder dataset by including different shapes and multiple objects.

The outline of this paper is as follows. 
In Section \ref{sec_mgn}, we introduce neural networks, MeshGraphNets method and the Navier-Stokes equations.
Next, in Section \ref{sec_numerical_tests}, we present the experiment setup and discuss the generalization of MeshGraphNets that have been trained on one obstacle shape to another obstacle shape.
Our work is summarized in Section \ref{sec_conclusion}.

\section{MeshGraphNets}
\label{sec_mgn}

We briefly recapitulate the MeshGraphNets (MGN) method introduced by Pfaff et al. in \cite{MGN}. MGNs are a deep learning approach for predicting the evolution of complex physics systems by applying graph neural networks to a mesh graph. The method is based on the idea of encoding the system state into a mesh graph, applying a graph neural network to this graph to predict the change of the system, and then applying this change to the simulation state. The method is autoregressive, meaning that the previous output can be used as the new input, allowing for the simulation to be rolled out to obtain the next states. The method is trained using a differentiable loss function and noise is added to the training data to force the MGN to correct any mistakes before they can accumulate in future states.

We first give an abstract definition of neural networks and then introduce the concept of MGNs for fluid dynamics simulations governed by the Navier-Stokes equations.

\subsection{Neural Networks}
An artificial neural network (ANN) is a parametrized, nonlinear function composition. By the
\textit{universal function approximation theorem} \cite{hornik1989multilayer},
arbitrary Borel measurable functions can be approximated by ANN. There are
several different formulations for ANN, which can be found in standard
references such as \cite{bishop2006pattern, goodfellow2016deep,
aggarwal2018neural, geron2019hands, chollet2018deep}. Following
\cite{hauser2018principles}, most ANN formulations can be unified.
An ANN $\mathcal{N}$ is a function from an \textit{input space}
$\mathbb{R}^{d_x}$ to an \textit{output space} $\mathbb{R}^{d_y}$, defined by a
composition of nonlinear functions $\bm{h}^{(l)}$, such that 
\begin{align} 
    \mathcal{N}:
    \mathbb{R}^{d_x} &\to \mathbb{R}^{d_y} \nonumber \\ \bm{x} &\mapsto
    \mathcal{N}(\bm{x}) = \bm{h}^{(l)} \circ \ldots \circ \bm{h}^{(0)} = \bm{y},
    \quad l = 1, \ldots, n_L.  
    \label{eq:ann} 
\end{align} 
Here, $\bm{x}$ denotes an \textit{input vector} of dimension $d_x$ and $\bm{y}$
an \textit{output vector} of dimension $d_y$. The nonlinear functions
$\bm{h}^{(l)}$ are called \textit{layers} and define an $l$-fold composition,
mapping input vectors to output vectors.
Consequently, the first
layer $\bm{h}^{(0)}$ is defined as the \textit{input layer} and the last layer
$\bm{h}^{(n_L)}$ as the \textit{output layer}, such that 
\begin{equation}
    \bm{h}^{(0)} = \bm{x} \in \mathbb{R}^{d_x}, \qquad \bm{h}^{(n_L)} = \bm{y}
    \in \mathbb{R}^{d_y}. 
    \label{eq:layer} 
\end{equation} 
The layers $\bm{h}^{(l)}$ between the
input and output layer, called \textit{hidden layers}, are defined as
\begin{equation} 
    \bm{h}^{(l)} \crowr \bm{h}_{\bullet}^{(l)} = \left\{h_{\bullet,
        \eta}^{(l)}, \; \eta = 1, \ldots,
    n_{u}\right\}, \qquad h_{\bullet, \eta}^{(l)} = 
    \varphi^{(l)} \circ \;
    \phi^{(l)}\left(\bm{W}^{(l)}_{\eta} \bullet \bm{h}^{(l-1)}\right),
    \label{eq:hidden} 
\end{equation} 
where $h_{\bullet, \eta}^{(l)}$ is the $\eta$-th \textit{neural unit} of the
$l$-th layer $\bm{h}_{\bullet}^{(l)}$ and $n_u$ is the \textit{total number of
neural units per layer}, while $\bullet$ denotes a product. Following the
notation in \cite{kenningtondifferential}, the symbol $\crowr$ denotes an
abbreviation of a tuple of mathematical objects $(\mathcal{O}_1, \mathcal{O}_2,
...)$, such that $\mathcal{O} \crowr (\mathcal{O}_1, \mathcal{O}_2, ...)$.
In \eqref{eq:hidden}, the details of type-specific layers
$\bm{h}_{\bullet}^{(l)}$ are gathered in general
layers $\bm{h}^{(l)}$ from \eqref{eq:ann}. The specification follows from the
$\bullet$-operator, which denotes the operation between the \textit{weight
vector} $\bm{W}_{\eta}^{(l)}$ of the $\eta$-th neural unit in the $l$-th layer
$\bm{h}_{\bullet}^{(l)}$ and the output of the preceding layer
$\bm{h}_{\bullet}^{(l-1)}$, where the bias term is absorbed
\cite{aggarwal2018neural}. If $\bullet$ is the
ordinary matrix multiplication $\bullet = \cdot$, then the layer
$\bm{h}_{\cdot}^{(l)}$ is called \textit{dense layer}. In the context
of GNNs, the choice of the operator as $\bullet = \oplus$, where $\oplus$ is a
permutation invariant aggregation operator, yields the so-called
\textit{message passing} network \cite{Bronstein2021Book}.

Furthermore, $\phi^{(l)}: \mathbb{R} \to \mathbb{R}$ is a nonlinear
\textit{activation function} and $\varphi^{(l)}$ is a function of the previous
layer, such that $\varphi^{(l)}: \bm{h}^{(l-1)} \mapsto
\varphi^{(l)}(\bm{h}^{(l-1)})$. If $\varphi^{(l)}$ is the identity function, the
layer $\bm{h}^{(l)}$ is called a \textit{feedforward layer}. All weight vectors
$\bm{W}_{\eta}^{(l)}$ of all layers $\bm{h}^{(l)}$ can be gathered in a single
expression, such that 
\begin{equation}
    \bm{\theta}=\left\{\bm{W}_{\eta}^{(l)}\right\}, 
    \label{eq:parameters}
\end{equation} 
where $\bm{\theta}$ inherits all parameters of the ANN $\mathcal{N}(\bm{x})$
from \eqref{eq:ann}. Consequently, the notation $\mathcal{N}(\bm{x};
\bm{\theta})$ emphasises the dependency of the outcome of an ANN on the input on
the one hand and the current realization of the weights on the other hand. The
specific combination of layers $\bm{h}_{\bullet}^{(l)}$ from
\eqref{eq:hidden}, neural units $h_{\bullet, \eta}^{(l)}$ and activation
functions $\phi^{(l)}$ from \eqref{eq:hidden} is called \textit{topology} of
the ANN $\mathcal{N}(\bm{x}; \bm{\theta})$. The weights $\bm{\theta}$ from
\eqref{eq:parameters} are typically found by gradient-based optimization with
respect to a task-specific $\textit{loss function}$ \cite{goodfellow2016deep}.

\subsection{Mesh Graph Networks}

In this section, we introduce MGN, a deep learning approach for predicting the evolution of complex physics systems by applying graph neural networks to a mesh graph. This is based on the paper \cite{MGN} by Pfaff et al. which is an extension of their previous paper on graph network based simulations \cite{MGN_pre_paper}. 
In the following and overall in this paper, we introduce and test MGNs for the Flow around the Cylinder benchmark (abbreviated as Cylinder Flow). However, we notice that 
our developments can be applied to other computational fluid dynamics numerical tests as well.

\subsubsection{Incompressible Navier-Stokes equations}
Let $\Omega\subset\mathbb{R}^2$ be
some domain with sufficiently smooth boundary $\partial\Omega$. Moreover,
let $(0,T)$ be the time interval with end time value $T>0$.
To generate training data, we model incompressible, viscous fluid flow with constant density and temperature by the Navier-Stokes equations
\begin{align*}
    \partial_t \bm{v} -\nabla_x \cdot \bm\sigma + (\bm{v} \cdot \nabla_x)\bm{v} &= 0 \quad\text{in } \Omega\times (0,T) , \\
    \nabla_x \cdot \bm{v} &= 0 \quad\text{in } \Omega\times (0,T), \\
    \bm{v} &= \bm{v}_D  \quad\text{on }\partial\Omega \times (0,T), \\
    \bm{v}(0) &= \bm{v}^0 \quad\text{in } \Omega \times \{0\},
\end{align*}
where for the stress $\bm\sigma$ we use the unsymmetric stress tensor 
\begin{align*}
    \bm\sigma := \bm\sigma \begin{pmatrix} \bm v \\ p \end{pmatrix} = -pI + \nu \nabla_x \bm{v}.   
\end{align*}
Plugging the definition of the unsymmetric stress tensor into the Navier-Stokes equations leads to the following formulation:
\begin{form}[Incompressible Navier-Stokes equations]
\label{form:navier_stokes}
Find the vector-valued velocity $\bm{v}: \Omega \times (0,T) \rightarrow \mathbb{R}^2$ and the scalar-valued pressure $p: \Omega \times (0,T) \rightarrow \mathbb{R}$ such that
\begin{align*}
    \partial_t \bm{v} + \nabla_x p - \nu \Delta_x \bm{v}  + (\bm{v} \cdot \nabla_x)\bm{v}  &= 0 \quad\text{in } \Omega\times (0,T) , \\
    \nabla_x \cdot \bm{v} &= 0 \quad\text{in } \Omega\times (0,T), \\
    \bm{v} &= \bm{v}_D  \quad\text{on }\partial\Omega \times (0,T), \\
    \bm{v}(0) &= \bm{v}^0 \quad\text{in } \Omega \times \{0\}.
\end{align*}
\end{form}
\noindent Here, $\nu > 0$ is the kinematic viscosity, $\bm{v}^0$ is the initial velocity and $\bm{v}_D$ is a possibly time-dependent Dirichlet boundary value.
In the following, we directly address the time-discrete problem. 
To this end, $\bm{s}_k=(\bm{v}_k,p_k)\colon \overline{\Omega}\mapsto \mathbb{R}^3$ denotes the system state (i.e. the velocity and pressure field) with domain $\Omega \subseteq \mathbb{R}^2$ at time $\Delta t \cdot k$ with the time step size $\Delta t > 0$ and the time step index $k \in \mathbb{N}$. Additionally, we use $\tilde{\bm{s}}_k=(\tilde{\bm{v}}_k,\tilde{p}_k)\colon \overline{\Omega}\mapsto \mathbb{R}^3$ to denote our prediction for $\bm{s}_k$. 
To create the dataset, we employ the finite element method using an \textit{incremental pressure correction scheme} (IPCS) \cite{fenicstutorial} which is based on Chorin's method \cite{chorinsmethod}.

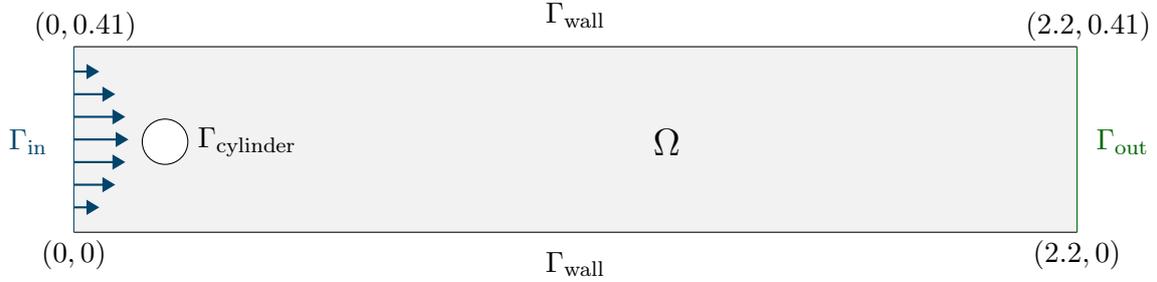
\begin{figure}[H]
    \begin{center}
    \begin{tikzpicture}[scale = 6, draw=black]
        \draw[draw=white, fill=black!5!white] (0,0) rectangle (2.2,0.41);
        \draw[draw=black] (0,0) -- (2.2,0);
        \draw[draw=black] (0,0.41) -- (2.2,0.41);
        \draw[draw=blue!60!black] (0,0) -- (0,0.41);
        \draw[draw=green!40!black] (2.2,0) -- (2.2,0.41);
        
        \draw[fill=white] (0.2,0.2) circle (0.05);
        \node (circ) at (0.38,0.2) {$\Gamma_{\text{cylinder}}$};
        
        \node (omega) at (1.3,0.2) {\Large{$\Omega$}};
        
        \node (in) at (-0.1,0.2) {{\color{blue!60!black}$\Gamma_{\text{in}}$}};
        \node (out) at (2.3,0.2) {{\color{green!40!black}$\Gamma_{\text{out}}$}};
        \node (wall-one) at (1.1,0.48) {$\Gamma_{\text{wall}}$};
        \node (wall-two) at (1.1,-0.07) {$\Gamma_{\text{wall}}$};
        
        \node (corner-one) at (0,-0.05) {$(0,0)$};
        \node (corner-two) at (0.025,0.455) {$(0,0.41)$};
        \node (corner-three) at (2.2,-0.05) {$(2.2,0)$};
        \node (corner-four) at (2.225,0.455) {$(2.2,0.41)$};
        
        \draw[draw=blue!60!black, -Triangle, thick] (0,0.055) -- (0.056,0.055);
        \draw[draw=blue!60!black, -Triangle, thick] (0,0.105) -- (0.091,0.105);
        \draw[draw=blue!60!black, -Triangle, thick] (0,0.155) -- (0.113,0.155);
        \draw[draw=blue!60!black, -Triangle, thick] (0,0.205) -- (0.12,0.205);
        \draw[draw=blue!60!black, -Triangle, thick] (0,0.255) -- (0.113,0.255);
        \draw[draw=blue!60!black, -Triangle, thick] (0,0.305) -- (0.091,0.305);
        \draw[draw=blue!60!black, -Triangle, thick] (0,0.355) -- (0.056,0.355);

    \end{tikzpicture}
    \caption{Domain of the Navier-Stokes benchmark problem (Cylinder Flow) \cite{dfg_benchmark}}
    \label{fig:channel_domain}
    \end{center}
\end{figure}

\subsubsection{MGN idea}
\begin{figure}[H]
      \centering
      
    \begin{tikzpicture}
    \node (A) at (0,0) {};
    
    \node at (4.5,2) {\includegraphics[width=1.75cm]{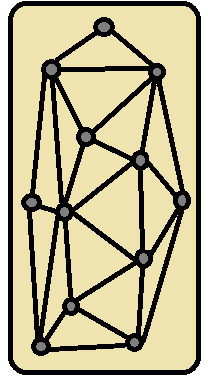}};
    \node (B) at (4.5,4) {$(G,\bm{s}_k)$};
    
    \draw (1, 0.5) rectangle (3, -0.5) node[midway] (encoder) {Encoder};
    \draw (6, 0.5) rectangle (8, -0.5) node[midway] (mgn) {MGN};
    \draw (11, 0.5) rectangle (13, -0.5) node[midway] (update) {Update};
    
    \draw[->] (0,0) -- node[above] {$\bm{s}_k$} (1,0);
    \draw[->] (3,0) -- node[above] {} (6,0);
    \draw[->] (8,0) -- node[above] {$MGN(G,\bm{s}_k)$} (11,0);
    \draw[->] (13,0) -- node[above] {$\tilde{\bm{s}}_{k+1}$} (14,0);
    \draw[->] (12,1.5) -- node[right] {$\bm{s}_k$} (12,0.5);
    \end{tikzpicture}
      \caption{High-level overview of how MGNs predict the next state given the current state $\bm{s}_k$.}
      \label{fig:mgn}
    \end{figure}
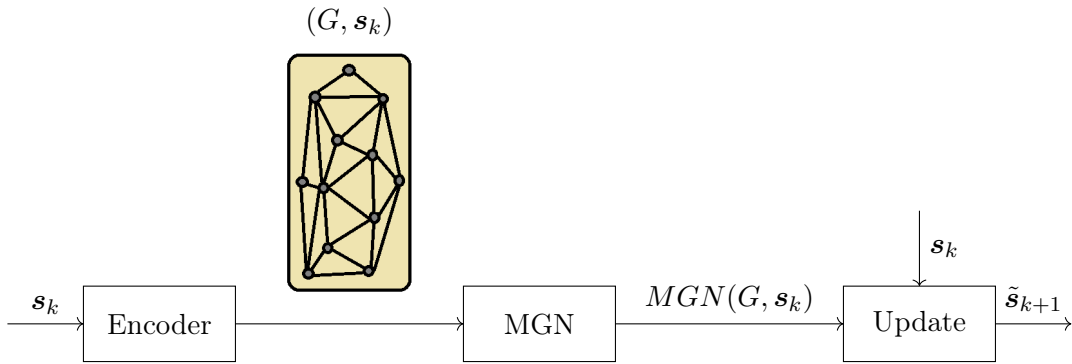

On a high level, the MGN approach can be explained as follows: First, we triangulate the domain to represent it as a mesh graph. 
We then encode the current system state $\bm{s}_k$ into the graph's nodes and edges. Next, we use a graph neural network to predict quantities which can be used to directly compute how $\bm{s}_k$ evolves to $\bm{s}_{k+1}$. In Cylinder Flow, we will use an MGN to directly predict the pressure field $p_{k+1}$, and the change in velocity $\bm{\delta v} = \bm{v}_{k+1} - \bm{v}_k$. Fig. \ref{fig:mgn} visualizes this process.

By construction, MGNs are autoregressive which means we can use the previous output as new input. Hence we can rollout a state $\bm{s}_k$ to obtain $\bm{s}_{k+1},\bm{s}_{k+2},\dots$. We will now discuss this on a more detailed level.

\subsubsection{System state encoding}
In this step, we want to encode the system state $\bm{s}_k$ into a graph.
As a first step, the domain $\overline{\Omega}$ has to be triangulated to obtain a mesh graph $G = (V, E)$ where $V \subseteq \mathbb{N}$ are the vertices and $E \subseteq \{\bm{e} \in \mathcal{P}(V) : |\bm{e}| = 2\}$ are the edges.

Next, we encode the position information into the graph: Let $\bm{e} = \{\nu_1,\nu_2\} \in E$ and $\bm{x}_{\nu_1},\bm{x}_{\nu_2} \in \overline{\Omega}$ be the domain coordinates associated with these two vertices. We assign the distance $\|\bm{x}_{\nu_1} - \bm{x}_{\nu_2}\|_2$ and the relative displacement $\bm{x}_{\nu_1} - \bm{x}_{\nu_2}$ as features to $\bm{e}$. It would also be possible to encode the absolute positions in the nodes. Experiments showed, however, that this is not a viable option \cite{MGN}. To better capture the structure of the mesh, the node type is added as a feature to each vertex. In the case of Cylinder Flow there are four types of nodes:
\begin{itemize}
    \item fluid nodes,
    \item wall (top, bottom, and cylinder) nodes,
    \item inflow nodes,
    \item outflow nodes.
\end{itemize}
Besides the node type we also add the values of the fields of $\bm{s}_k$ needed to predict $\bm{s}_{k+1}$ as features to each vertex. For example, in the case of Cylinder Flow, each vertex $\nu \in V$ with associated coordinates $\bm{x} \in \overline{\Omega}$ gets the value of the velocity field $\bm{v}_k$ at position $\bm{x}$. Though pressure is also part of $\bm{s}_{k}$, it is not encoded as $\bm{v}_k$ already determines the pressure values hence they are not needed to predict $\bm{s}_{k+1}$.
The resulting graph and its associated features now approximately model $\bm{s}_k$.

\subsubsection{Graph processing}
In this step, we use the graph-encoding of the previous step to predict $\bm{s}_{k+1}$. For the incompressible Navier-Stokes equations, we directly predict the pressure $p_{k+1}$ and the change in velocity $\bm{\delta v} = \bm{v}_{k+1}-\bm{v}_k$. The processing is done in three different steps. 
\begin{enumerate}
    \item First, to prepare the features for processing, we use a feedforward-network to encode the edge features and another feedforward-network to encode the vertex features at each edge/vertex. We say that the features are embedded into latent space. In theory, this step is not needed. However, in practice without this step, the network would not perform well. This is because in the next step, these features will iteratively be updated to finally arrive at the prediction for the quantity change. If we do not encode the features in this first step, the information captured in these intermediate states is always restricted to the dimension of the input features. Therefore, this step can be viewed as widening the information bottleneck.
    \item Next, we process the previously encoded features. First, we apply a
neural network $\mathcal{N}^M$ to all edges to update their edge features. Let $\bm{e} = \{\nu_1,\nu_2\} \in E$ and $\mathcal{F}(\nu_1),\mathcal{F}(\nu_2),\mathcal{F}(\bm{e})$ be the features associated with this edge/nodes. The updated edge features $\mathcal{F}^{\prime}(\bm{e})$ are given as follows:
    \begin{equation}
        \mathcal{F}^{\prime}(\bm{e}) = \mathcal{N}^M(\mathcal{F}(\bm{e}),\mathcal{F}(\nu_1),\mathcal{F}(\nu_2)).
    \end{equation}
    Using $\mathcal{F}^{\prime}(\bm{e})$, we update the node-features using
another neural network $\mathcal{N}^V$. Let $\nu \in V$ and $N_\nu \subseteq E$ be the set of edges connected to $\nu$. Again, we denote their associated features with $\mathcal{F}(\nu)$ and $\mathcal{F}^{\prime}(\bm{e})$ for $\bm{e} \in N_\nu$. The updated features $\mathcal{F}^{\prime}(\nu)$ are defined as follows:
    \begin{equation}
        \mathcal{F}^{\prime}(\nu) = \mathcal{N}^V \left (\mathcal{F}(\nu),\sum\limits_{\bm{e} \in N_\nu} \mathcal{F}^{\prime}(\bm{e}) \right ) .
    \end{equation}
    After this, we simply replace $\mathcal{F}(\nu)$ by
$\mathcal{F}^{\prime}(\nu)$ and $\mathcal{F}(\bm{e})$ by
$\mathcal{F}^{\prime}(\bm{e})$ for each vertex/edge. This is called one
\textit{message passing block} (MPB). In total, we apply $L$ identical message
passing blocks to the features (i.e. $\mathcal{N}^M,\mathcal{N}^V$ are fixed). 
    
    Note that within each message passing block, information can flow only from
one vertex to all its neighbors, because its information first passes to its
edges by $\mathcal{N}^M$ and then to its neighbors by $\mathcal{N}^V$. Hence with $L$ MPBs only vertices with a distance of less than $L$ steps can be reached which means that even though in theory nodes can influence each other no matter their distance, this is not captured by this processing step.

    \item In this last step, we apply a final feedforward-net (called the decoder) to the latent features of each vertex obtained after the last MPB step. This output will then be used to directly update the system state as will be described in the following.
    
\end{enumerate}

\subsubsection{Updating the system state}
Assume that $\bm{q}_{k+1}: V \mapsto \mathbb{R}^3$ is the output of our MGN with input $\bm{s}_{k}$.
We interpret $\bm{q}_{k+1}$ as the derivative of the velocity field $\bm{v}_k$ and the pressure $p_{k+1}$, i.e. $q_{k+1} = (\bm{q}_{1,k+1},q_{2,k+1}) \approx (\bm{\partial_t} \bm{v}_k,p_{k+1})$ where $\bm{\partial_t} \bm{v}_k$ denotes the time derivative of the velocity field at time $\Delta t \cdot k$. Our prediction $\tilde{\bm{s}}_{k+1}$ at node $\nu \in V$ with coordinates $\bm{x} \in \overline{\Omega}$ is then given by 
\begin{equation}
    \tilde{\bm{s}}_{k+1}(\bm{x}) = (\bm{v}_k(\bm{x}) + \Delta t \cdot \bm{q}_{1,k+1}(\nu), q_{2,k+1}(\nu)).
\end{equation}

\subsubsection{Training MGNs}
\label{subsec:trainin_mgn}
In this section, we discuss the training details of MGNs. First, note that each step as detailed in the previous section is differentiable. Hence, the predicted quantities at each vertex are differentiable with respect to the parameters of all neural networks that were used. Therefore, we can define a differentiable loss $\mathcal{L}$ at each node $\nu \in V$ with coordinates $\bm{x} \in \overline{\Omega}$ by taking the mean-squared-error of the MGN prediction and the ground truth field values. Hence, the loss $\mathcal{L}$ is
\begin{equation}
\label{eq:mgn_loss}
  \mathcal{L}(\nu) = \frac{1}{3} \|\tilde{\bm{s}}_{k+1}(\bm{x}) - \bm{s}_{k+1}(\bm{x})\|_2^2.
\end{equation}
The MGN can then be trained using gradient descent on $\mathcal{L}$ averaged on each vertex. Note that this requires access to the ground truth values $\bm{s}_k$ at the vertices or approximations of these which might not always be efficiently available.

Though this training setup works in theory, experiments \cite{MGN} showed that this does not suffice to predict states far in the future and not just the next time step. This problem is addressed by adding noise to the training data and thus forcing the MGN to correct any mistakes before they can accumulate in future states. For all simulations in the training data, at every time step $k$ and at each vertex $\nu \in V$ with coordinates $\bm{x} \in \overline{\Omega}$, we independently sample normally distributed noise $\bm{\varepsilon} = (\varepsilon_1,\varepsilon_2)$ with $\varepsilon_1, \varepsilon_2 \sim \mathcal{N}(0,\sigma)$ with zero mean and a fixed variance $\sigma \in \mathbb{R}^+$. This noise is then added to the velocity field $\bm{v}_k(x)$ to obtain the noisy velocity field $\bm{v}^{noisy}_k(x) = \bm{v}_k(x) + \bm{\varepsilon}$. Then, instead of $\bm{v}_k$, the noisy velocity field $\bm{v}^{noisy}_k$ is fed as the input to the MGN whose next state prediction we then denote by $\tilde{\bm{s}}^{noisy}_{k+1}$ which is then used in Equation (\ref{eq:mgn_loss}) instead of $\tilde{\bm{s}}_{k+1}$. The target $\bm{s}_{k+1}$ remains unchanged. This modification allows the MGN to learn how to denoise the system state thereby learning how to reduce error accumulation as shown experimentally by DeepMind \cite{MGN}.

\section{Generalization Capabilities of MeshGraphNets and Numerical Simulations}
\label{sec_numerical_tests}
In this section, we will present experimental results on the generalization of MGNs to unseen datasets. First, in Section \ref{sec:exp_setup} we will describe our experiment setup. Then, in Section \ref{sec:exp_gen} we will present the datasets of the generalization experiments, followed by the results in Section \ref{sec:exp_gen_results} and the runtimes in Section \ref{sec:exp_runtimes}.
The code used for these experiments is publicly available \cite{github_implementation}. 

\subsection{Setup of the experiments}
\label{sec:exp_setup}
In this section, we will describe the exact setup for the MGNs used for the generalization experiments. We will also describe which evaluation metrics for the MGNs were used for the subsequent experiments.

\subsubsection{Implementation}
\label{subsec:implementation}
For the generalization experiments, we use the implementation contained in NVIDIA's Modulus, a machine learning framework for physics \cite{NVIDIA_SimNet, NVIDIA_Modulus}.
More specifically, the exact implementation contained in \textbf{examples/cfd/vortex\_shedding\_mgn}\footnote{\url{https://github.com/NVIDIA/modulus/tree/main/examples/cfd/vortex_shedding_mgn}} from NVIDIA Modulus \cite{NVIDIA_Modulus} was used for experimentation with modifications only affecting logging and the learning rate schedule which is now only updated after every epoch such that the learning rate in epoch $l$ is $\eta_1 \gamma^{l-1}$ for some initial learning rate $\eta_1$ and learning rate decay $\gamma$.

\subsubsection{MGN architecture details}
We highlight some notable architectural choices of the MGNs.
\begin{itemize}
    \item Input and target normalization: The training data set is used to compute the mean and variance of all edge features and of the velocity and pressure for the node features. The features of all inputs and targets are then normalized using their corresponding mean and variance.
    \item Residual connections: The MGN will be extremely deep, therefore residual connections are added such that the input of each MBP is directly added to its output.
    \item Layer normalization: Since we will be using a batch size of one but still want to normalize the hidden-layer outputs to reduce the covariate shift (i.e. gradients in layer $i$ tend to highly depend on the outputs of layer $i-1$), we employ layer normalization \cite{ba2016layer} at the output of the edge/node encoder and after each MPB (before the residual connection). Layer normalization transforms an output $x \in \mathbb{R}^n$ as follows:
    \begin{equation}
        y = \frac{x - \mathbb{E}[x]}{\sqrt{\mathrm{Var}[x]+\epsilon}} \cdot \alpha + \beta
    \end{equation}
    where $\mathbb{E}[x]$ is subtracted componentwise and
    \begin{equation}
        \mathbb{E}[x] = \frac{1}{n} \sum\limits_{i=1}^n x_i, \ \ \ \mathrm{Var}[x] = \frac{1}{n} \sum\limits_{i=1}^n (x_i - \mathbb{E}[x])^2
    \end{equation}
    where $\alpha,\beta \in \mathbb{R}$ are learnable parameters and $\epsilon > 0$ for numerical stability. We used $\epsilon = 10^{-5}$.
\end{itemize}

\subsubsection{Hyperparameters}
\label{subsec:hyperparams}
The exact hyperparameters from \textbf{examples/cfd/vortex\_shedding\_mgn}\footnote{\url{https://github.com/NVIDIA/modulus/tree/main/examples/cfd/vortex_shedding_mgn}} from NVIDIA Modulus \cite{NVIDIA_Modulus} were used for the subsequent experiments. The only exception is the learning rate decay factor $\gamma$ since we used a slightly different implementation of the learning rate decay schedule as described in Subsection \ref{subsec:implementation}. Table \ref{tab:hyperparams} lists the hyperparameters.

\begin{table}[H]
    \centering
    
    \begin{tabular}{|c|c|c|}
        \hline
        \textbf{Parameter} & \textbf{Value} & \textbf{Comment} \\
        \hline
        Epochs & 25 & Iterate entire dataset\\
       & & per epoch\\
        Batch size & 1 &  \\
        Initial learning rate $\eta_1$ & 0.0001 &  \\
        Learning rate decay $\gamma$ & 0.82540 & $\eta_l = \eta_1 \gamma^{l-1}$ at epoch $l$. \\
        Message passing blocks $L$ & 15 &  \\
        Encoders/Decoder/Processors hidden layers & 2 &  \\ 
        Encoders/Decoder/Processors hidden dim & 128 &  \\
        Node embedding size & 128 &  \\
        Edge embedding size & 128 &  \\
        Activation function & ReLU &  \\
        Optimizer & Adam &   \\
        Parameter initialization & PyTorch &  PyTorch v1.11.0 \\ 
        & default &  \\ 
        Training noise & $\mathcal{N}(0,0.02)$ & \\
        \hline
    \end{tabular}

    \caption{Hyperparameters used for the experiments.}
    \label{tab:hyperparams}
\end{table}

\subsubsection{Evaluation criteria}
\label{subsec:err_metric}
For each experiment, we use at least 10\% of the total dataset for evaluation. The exact evaluation datasets will be mentioned in the corresponding experiment. Let $K \in \mathbb{N}$ denote the number of simulations, let $J \in \mathbb{N}$ denote the number of time steps in each simulation, and let $I_k \in \mathbb{N}$ denote the number of mesh nodes in the $k$-th simulation with $k \leq K$ of the evaluation dataset. For evaluation, we measure eight quantities. The first four are derived from the root-mean-squared-error (RMSE) of the velocity. Let $v_{i,j,k} \in \mathbb{R}^2$ denote the velocity vector at mesh node $i \leq I_k$ and time step $j \leq J$ in the $k$-th  ($k \leq K$) simulation of the evaluation dataset obtained by performing the IPCS (more details on the dataset generation in the next section). Furthermore, we use $\tilde{v}_{i,j,k}$ to denote the MGN velocity prediction from $v_{i,j-1,k}$ if $j > 1$ and $v_{i,1,k}$ if $j=1$. Lastly, by $\hat{v}_{i,j,k}$ we denote the MGN velocity prediction from $\hat{v}_{i,j-1,k}$ if $j > 1$ and $v_{i,1,k}$ if $j=1$. From which MGN the predictions come will be specified every time one of the following error quantities is reported.
\begin{itemize}
    \item \textbf{Velocity 1-step error} \bm{$\epsilon_1^{(v)}$}: This error is the RMSE for all the next state velocity predictions of all timesteps of all simulations which is given by
        \begin{equation}
            \epsilon_{1}^{(v)} = \sqrt{\sum \limits_{k=1}^K \sum \limits_{j=1}^J \sum \limits_{i=1}^{I_k} \frac{||v_{i,j,k} - \tilde{v}_{i,j,k}||_2^2}{2I_k \cdot J \cdot K}}.
        \end{equation}
    
    \item \textbf{Velocity 50-steps error} \bm{$\epsilon_{50}^{(v)}$}: Each simulation is rolled out for 50 timesteps. The RMSE is then taken for the rollout velocity at each of the first 50 time steps compared to the IPCS result for all simulations. This given by
    \begin{equation}
            \epsilon_{50}^{(v)} = \sqrt{\sum \limits_{k=1}^K \sum \limits_{j=1}^{50} \sum \limits_{i=1}^{I_k} \frac{||v_{i,j,k} - \hat{v}_{i,j,k}||_2^2}{2I_k \cdot 50 \cdot K}}.
        \end{equation}
    \item \textbf{Velocity all-steps error} \bm{$\epsilon_{all}^{(v)}$}: Each simulation is fully rolled out. The RMSE then compares the rollout velocity prediction at each time step to the IPCS  velocity for all simulations. Formally, this is
    \begin{equation}
            \epsilon_{all}^{(v)} = \sqrt{\sum \limits_{k=1}^K \sum \limits_{j=1}^J \sum \limits_{i=1}^{I_k} \frac{||v_{i,j,k} - \hat{v}_{i,j,k}||_2^2}{2I_k \cdot J \cdot K}}.
    \end{equation}

    \item \textbf{Velocity all-steps median} \bm{$\epsilon_{all,median}^{(v)}$}: We first calculate the all-steps RMSE for each individual simulation $k \leq K$ as
     \begin{equation}
        \epsilon_{all,k}^{(v)} = \sqrt{ \sum \limits_{j=1}^J \sum \limits_{i=1}^{I_k} \frac{||v_{i,j,k} - \hat{v}_{i,j,k}||_2^2}{2I_k \cdot J}}.
    \end{equation}
    The error $\epsilon_{all,median}^{(v)}$ is then given by the median of $(\epsilon_{all,1}^{(v)},\dots,\epsilon_{all,K}^{(v)})$.
\end{itemize}

 We ran each experiment with three different seeds. For each of the above error quantities, we report the mean and the maximum deviation from the mean across all three seeds. For example, we would report the three values $\{0,3,12\}$ as $5 \pm 7$. 

 The remaining four error quantities \bm{$\epsilon_1^{(p)}$},\bm{$\epsilon_{50}^{(p)}$},\bm{$\epsilon_{all}^{(p)}$}, and \bm{$ \epsilon_{all,median}^{(p)}$} come from doing the same with the pressure predictions.

\subsection{Generalization experiments: Datasets}
\label{sec:exp_gen}
The experiments in this section have the aim of expanding DeepMind's MGN generalization experiments. DeepMind claims that their MGN has strong generalization capabilities due to using a relative encoding on the mesh graphs, however, they report no experimental evidence for this for their Cylinder Flow based dataset at all \cite{MGN}. Additionally, in their Airfoil experiments of compressible Navier-Stokes equations, which is a related problem to Cylinder Flow, they only test the MGN by using values for the system parameters (such as the inflow speed) that are slightly out of the training distribution. What they do not do is test the MGN on qualitatively different meshes such as one where a different shape other than a circle is used.

Therefore, in this section, we apply a trained MGN to similar but qualitatively different datasets than it was trained on in which the cylinder may be stretched or squeezed (\textbf{cylinder\_stretch)}, the cylinder might be replaced with a triangle or rectangle (\textbf{cylinder\_tri\_quad}), there could even be multiple cylinders (\textbf{2cylinders}), or all modifications mixed together (\textbf{mixed\_all}) or simply no modifications (\textbf{standard\_cylinder}). 

The datasets consist of approximations of solutions (velocity $\bm{v}$ and pressure $p$) to the Cylinder Flow problem. The datasets are created by varying the inflow profile or applying changes to the cylinder. 
Whenever possible, the distributions for the parameters we used were reverse-engineered from Google Deepmind's dataset \cite{MGN} assuming that uniform distributions were used. One notable difference to the DFG\footnote{DFG stands for German Research Foundation} 2D-2 benchmark problem \cite{dfg_benchmark} is that the pipe is a bit shorter ($1.6$ instead of $2.2$ length) to speed up computation. Other than that, the range of parameters includes this benchmark problem. We note that the mean values for certain quantities such as the $x$-position of the cylinder are not always equal to the one in the benchmark problem as we had to include some wiggle room to avoid unphysical situations (e.g. the cylinder intersects the inflow). 

The domain $\Omega$ for a single simulation is always of the form $\Omega = [(0,1.6) \times (0,0.41)] \backslash \overline{\mathcal{C}} \subseteq \mathbb{R}^2$, where $\mathcal{C} \subseteq \mathbb{R}^2$ is distributed depending on the dataset.
\begin{itemize}
    \item \textbf{standard\_cylinder}: In this dataset $\mathcal{C}$ is always a circle with midpoint $(x,y)$ where $x \sim \mathcal{U}[0.15,0.5]$ and $y \sim \mathcal{U}[0.1,0.3]$ are sampled independently. The radius is sampled from $r \sim \mathcal{U}[0.02,0.08]$.
    \item \textbf{cylinder\_stretch}: Same as in \textbf{standard\_cylinder} except that the circle may now be an ellipse with height $h \sim  \mathcal{U}[0.02,0.08]$ and width $w \sim  \mathcal{U}[0.02,0.08]$.
    \item \textbf{cylinder\_tri\_quad}: There is a $\frac{1}{3}$ chance that a circle as in \textbf{standard\_cylinder} is chosen. There is another $\frac{1}{3}$ chance that a square with a midpoint sampled the same way as the ellipse's and side length $s \sim \mathcal{U}[0.02 \sqrt{2}, 0.08 \sqrt{2}]$ and a random orientation with a uniformly sampled angle is chosen. And another $\frac{1}{3}$ chance that an equilateral triangle with side length $s \sim \mathcal{U}[0.078,0.182]$ and a random orientation with a uniformly sampled angle and midpoint sampled the same way as the ellipse's is chosen as $\mathcal{C}$.
    \item \textbf{2cylinders}: $\mathcal{C} = \mathcal{C}_1 \cup \mathcal{C}_2$. Here $\mathcal{C}_1$ is an object as in \textbf{standard\_cylinder}. With a $\frac{1}{2}$ chance, we choose $\mathcal{C}_2 = \emptyset$ and otherwise $\mathcal{C}_2$ is sampled the same as $\mathcal{C}_1$ except that the mean-midpoint is now $(0.825,0.2)$ instead of $(0.325,0.2)$.
    \item \textbf{mixed\_all}: $\mathcal{C} = \mathcal{C}_1 \cup \mathcal{C}_2$. Here $\mathcal{C}_1$ is an object as in \textbf{cylinder\_tri\_quad} except that each object is randomly stretched or squeezed (a circle is stretched as in \textbf{cylinder\_stretch}, a square samples width and height independently as its side length, and a triangle stretches around its midpoint with the factor $s \sim \mathcal{U}[0.7,1.3]$ in the $x$-direction and $\frac{1}{s}$ in the $y$-direction). With a $\frac{1}{4}$ chance $\mathcal{C}_2$ is sampled the same as $\mathcal{C}_1$ except that the midpoint is now shifted by $0.5$ in the $x$-direction, i.e. it is now $(0.825,0.2)$. However, with a $\frac{3}{4}$ chance we choose $\mathcal{C}_2 = \emptyset$.
\end{itemize}

\begin{figure}[H]
\subfloat[][\textbf{standard\_cylinder}]
{\includegraphics[width=0.45\textwidth]{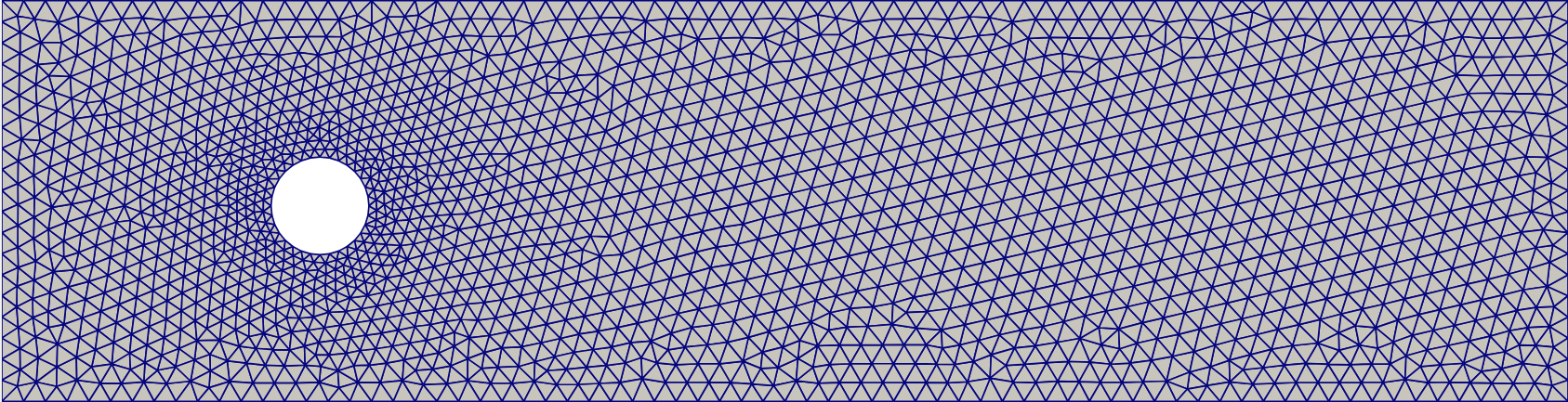}}
\hfill
\subfloat[][\textbf{cylinder\_stretch}]
{\includegraphics[width=0.45\textwidth]{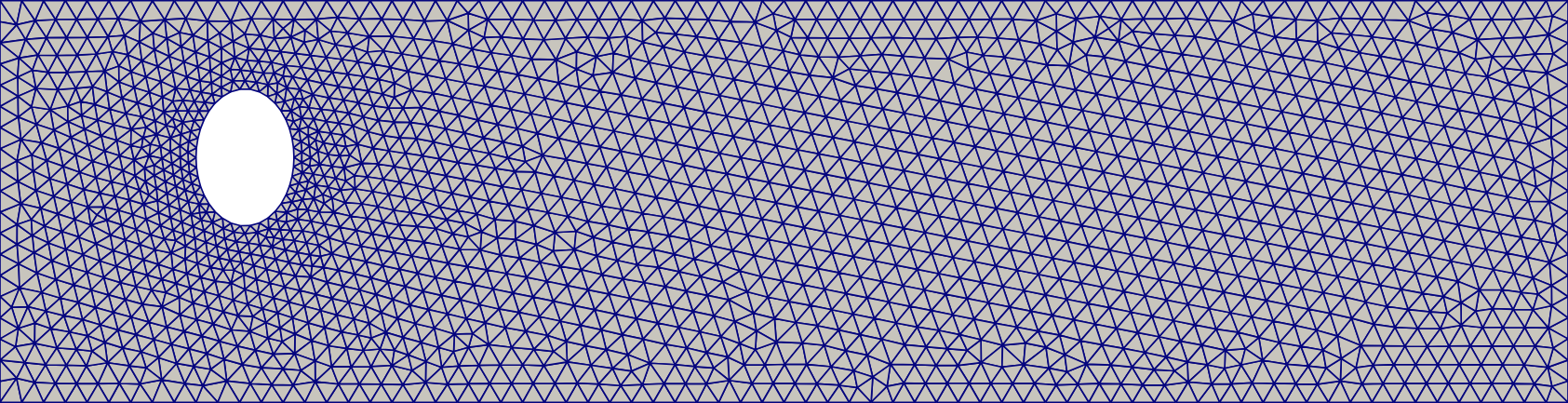}}

\subfloat[][\textbf{cylinder\_tri\_quad}]
{\includegraphics[width=0.45\textwidth]{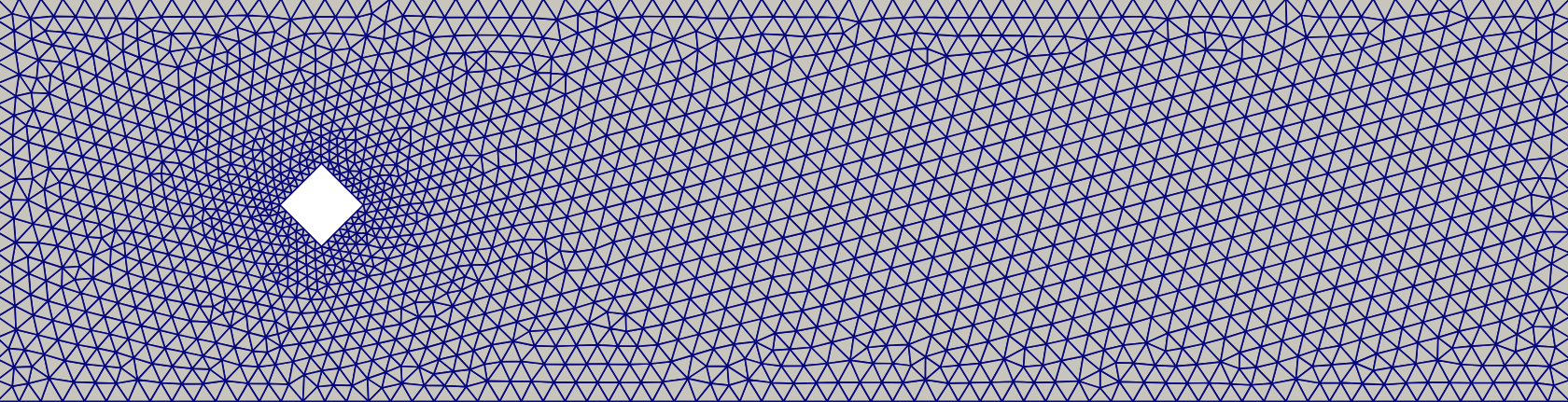}}
\hfill
\subfloat[][\textbf{2cylinders}]
{\includegraphics[width=0.45\textwidth]{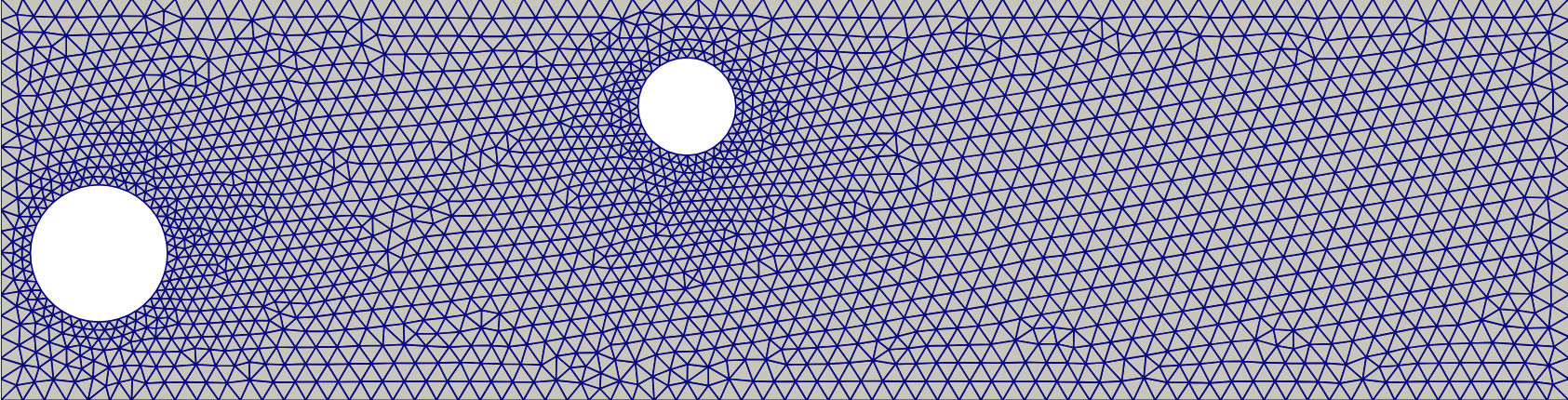}}

\centering
\subfloat[][\textbf{mixed\_all}]
{\includegraphics[width=0.45\textwidth]{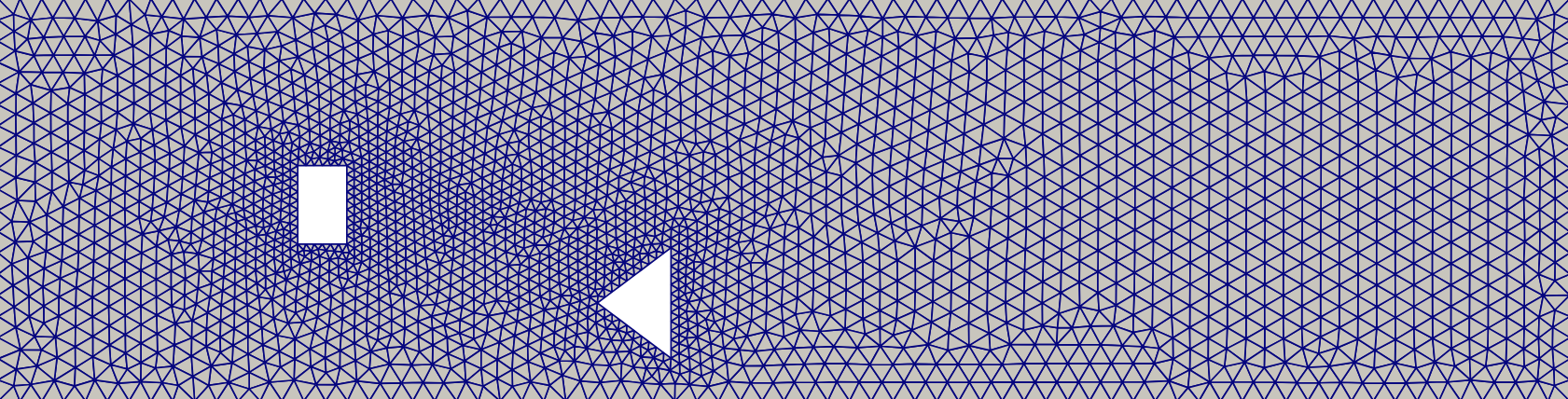}}
\caption{One example mesh from each of the five datasets.}
\label{fig:meshes}
\end{figure}

All of the above-described domains are triangulated with pygmsh \cite{pygmsh} using the parameters ${mesh\_size=0.0225}$ for the domain and ${mesh\_size=0.0098}$ for the geometric objects to have a finer mesh around the region of interest. We chose these parameters to get to a similar mesh size that DeepMind used for their dataset.
Furthermore, to avoid extreme edge cases, we exclude meshes where the rectangular bounding box of the object(s) come(s) within a distance of less than $0.02$ to the boundary $\partial \Omega$.
For \textbf{standard\_cylinder} with the circle being centered at $(0.325,0.2)$ and $r=0.05$ we obtain a mesh with $2069$ vertices and $3924$ triangles/cells. For the slightly elongated pipe of width $2.2$ and circle centered at $(0.2,0.2)$ we get $2582$ vertices and $4898$ cells. Note that we do not use this elongated mesh for the experiments. We will only use it later to compare quantities of interest with the benchmark results to verify our computation. Fig. \ref{fig:meshes} shows one sample mesh for each mesh set.

The inflow profile is determined the same way for each mesh in each dataset. First, we sample $U \sim \mathcal{U}[0.25,2.25]$. The inflow profile is then given by:
\begin{equation}
    \bm{v}(x,y,t) = \left (U \cdot \frac{4y(0.41 -y)}{0.41^2},0 \right ),\ \ (x,y,t) \in (\{0\} \times [0,0.41]) \times I.
\end{equation}
\\ 
For a given mesh, the solution velocity $\bm{v}$ and pressure $p$ is approximated using the IPCS.
FEniCS \cite{fenicsproject} was used as the solver using inf-sup stable quadratic elements for velocity and linear elements for the pressure.
Each simulation within a dataset contains the velocity and pressure values at each vertex in the mesh that was used for the approximation from $t=0.01s$ to $t=3s$ in $0.01s$ intervals, i.e. $t \in \{m,2m,\dots,300m\}$ with $m = 1/100s$. As the initial velocity for the solver, we simply used $\bm{v},p\equiv 0$. Since this violates the boundary conditions, we exclude $t=0s$ and begin with $t=1/100s$ since at that time the boundary conditions are approximately satisfied. Even though the intervals are of size $0.01$, the solver used an adaptive $\Delta t$ under the hood that scales with the inflow peak. We used $\Delta t = 0.00025$ for an inflow peak of $1.25$. In total, each dataset contains $440$ simulations where $400$ are used for training, and $40$ for evaluation.

The above-described datasets can be generated by following the instructions provided in \textit{README.md} of the codebase \cite{github_implementation}. As mentioned in the introduction, these datasets combined, constitute a new benchmark for fluid simulations that extends DeepMind's Cylinder Flow based dataset \cite{MGN}.

\subsubsection{Comparison to DFG 2D-2 benchmark} ~\\
\label{sec:qoi_comparison}
To verify our experiment setup for generating datasets, we computed some quantities of interest and compared them to the Navier-Stokes 2D-2 benchmark results by Schäfer and Turek \cite{dfg_benchmark}. For this, we simulated the flow on the pipe with width $2.2$ and circle center at $(0.2,0.2)$ for $30$ seconds to make sure that the flow was fully developed. We considered the interval $[29.9295s,29.9625s]$ (the last interval between two maxima of the lift coefficient) for our measurements.
We measured the maximum value of the drag coefficient $c_D$ and the lift coefficient $c_L$ within this interval and the pressure difference $\Delta p$ at the interval-midpoint $t_{mid}=29.9460s$. We ran this computation on a system with six Intel(R) Core(TM) i5-9600K @ 3.70GHz CPUs with 32 GB of RAM. Table \ref{tab:qoi} shows our results.

\begin{table}[H]
    \centering
    \begin{tabular}{@{}lccc@{}}
        \toprule
        \addlinespace
        \textbf{Quantity} & \textbf{Results} & \textbf{Benchmark results} \\
        \addlinespace
        \midrule
        \addlinespace
        $St$ & $3.0303$ &  $[0.2950, 0.3050]$\\ 
        max $c_D$ & $3.1668$ & $[3.2200, 3.2400]$ \\ 
        max $c_L$& $1.0210$ & $[0.9900, 1.0100]$ \\ 
        $\Delta p$ & $2.5017$  & $[2.4600, 2.5000]$\\ 
        \addlinespace
        \bottomrule
    \end{tabular}

    \caption{Quantities of interest for DFG 2D-2 benchmark.}
    \label{tab:qoi}
\end{table}
We note that our values slightly deviate from the benchmark results but are still deemed acceptable.

\subsection{Generalization experiments: Results}
\label{sec:exp_gen_results}
We tested the generalization capabilities of the MGNs. For this, we tested each MGN on every dataset. E.g. train on \textbf{cylinder\_stretch} and evaluate on \textbf{2cylinders} and do this for every possible dataset-pair.
Table \ref{tab:rmse_cylinder} (resp. \ref{tab:rmse_stretch}, \ref{tab:rmse_quad}, \ref{tab:rmse_2}, \ref{tab:rmse_mixed}) shows the results if an MGN was trained on all of the five datasets and evaluated on \textbf{standard\_cylinder} (resp. \textbf{cylinder\_stretch, cylinder\_tri\_quad, 2cylinders, mixed\_all}) in terms of the pressure and velocity RMSE as described in Section \ref{subsec:err_metric}. Additionally, these tables also contain the RMSEs for newly initialized MGNs that were not trained to act as a benchmark for comparison. All reported velocity errors have been rounded to five decimal places and all pressure errors have been rounded to four decimal places.

\begin{table}[H]
\centering
\begin{minipage}{\linewidth}
\centering
\caption*{Velocity RMSE $\times 10^{-3}$}
\scalebox{0.9}{
\begin{tabular}{@{}lcccc@{}}
\toprule
\addlinespace
\textbf{Train Dataset} & \textbf{1-step} & \textbf{50-steps} & \textbf{all-steps} & \textbf{all-steps median} \\
\addlinespace
\midrule
\addlinespace 
standard\_cylinder & \bm{$2.32 \pm 0.02$} & \bm{$32.93 \pm 5.15$} & \bm{$89.19 \pm 17.2$} & \bm{$24.22 \pm 4.01$} \\
cylinder\_stretch & $2.56 \pm 0.01$ & $40.0 \pm 3.67$ & $100.52 \pm 14.05$ & $33.97 \pm 0.64$ \\
cylinder\_tri\_quad & $2.72 \pm 0.01$ & $36.46 \pm 2.39$ & $94.84 \pm 8.77$ & $34.97 \pm 3.58$ \\
2cylinders & $2.52 \pm 0.03$ & $35.22 \pm 4.12$ & $89.3 \pm 13.22$ & $35.03 \pm 3.44$ \\
mixed\_all & $2.98 \pm 0.01$ & $40.72 \pm 1.86$ & $115.44 \pm 11.12$ & $39.08 \pm 0.9$ \\
\textit{None} & $32.29 \pm 3.05$ & $473.44 \pm 214.41$ & $1398.48 \pm 618.28$  & $1369.2 \pm 635.18$\\
\addlinespace
\bottomrule
\end{tabular}}
\end{minipage}

\begin{minipage}{\linewidth}
\centering
\caption*{Pressure RMSE $\times 10^{-2}$}
\scalebox{0.9}{
\begin{tabular}{@{}lcccc@{}}
\toprule
\addlinespace
\textbf{Train Dataset} & \textbf{1-step} & \textbf{50-steps} & \textbf{all-steps} & \textbf{all-steps median} \\
\addlinespace
\midrule
\addlinespace
standard\_cylinder & \bm{$6.54 \pm 0.08$} & \bm{$5.69 \pm 0.3$} & \bm{$10.3 \pm 0.42$} & \bm{$1.77 \pm 0.2$}\\
cylinder\_stretch & $7.31 \pm 0.06$ & $7.36 \pm 0.14$ & $11.71 \pm 1.48$ & $2.03 \pm 0.14$\\
cylinder\_tri\_quad & $9.39 \pm 0.5$ & $9.57 \pm 0.7$ & $12.42 \pm 1.0$& $2.36 \pm 0.25$\\
2cylinders & $11.58 \pm 0.94$ & $12.45 \pm 0.99$ & $14.27 \pm 0.86$ &  $6.34 \pm 1.06$ \\
mixed\_all & $12.25 \pm 0.82$ & $13.78 \pm 0.94$ & $16.6 \pm 0.63$ & $3.99 \pm 0.65$\\
\textit{None} & $38.84 \pm 3.83$ & $36.2 \pm 3.89$ & $38.34 \pm 3.09$& $20.77 \pm 6.11$ \\
\addlinespace
\bottomrule
\end{tabular}}
\end{minipage}

\caption{RMSE if evaluated on \textbf{standard\_cylinder}. \textit{None} means that a newly initialized network without any training was evaluated. The lowest RMSE of each column has been marked.}
\label{tab:rmse_cylinder}

\end{table}

\begin{table}[H]
\centering

\begin{minipage}{\linewidth}
\centering
\caption*{Velocity RMSE $\times 10^{-3}$}
\scalebox{0.9}{
\begin{tabular}{@{}lcccc@{}}
\toprule
\addlinespace
\textbf{Train Dataset} & \textbf{1-step} & \textbf{50-steps} & \textbf{all-steps}& \textbf{all-steps median} \\
\addlinespace
\midrule
\addlinespace
standard\_cylinder & $7.33 \pm 0.15$ & $54.4 \pm 9.12$ & \bm{$199.34 \pm 10.76$} &$55.02 \pm 12.11$ \\ 
cylinder\_stretch & \bm{$3.52 \pm 0.03$} & \bm{$43.1 \pm 3.22$} & $221.25 \pm 21.11$ & $42.16 \pm 7.85$\\ 
cylinder\_tri\_quad & $4.5 \pm 0.05$ & $53.38 \pm 7.16$ & $217.1 \pm 30.32$& \bm{$40.71 \pm 6.3$}\\
2cylinders & $6.15 \pm 0.12$ & $85.71 \pm 7.76$ & $265.68 \pm 6.82$  & $73.28 \pm 7.94$ \\
mixed\_all & $3.91 \pm 0.03$ & $56.36 \pm 6.82$ & $210.39 \pm 13.78$ & $42.71 \pm 5.38$ \\
\textit{None} & $49.51 \pm 3.23$ & $903.62 \pm 552.82$ & $3111.27 \pm 2347.83$ & $3119.95 \pm 2352.9$ \\
\addlinespace
\bottomrule
\end{tabular}}
\end{minipage}

\begin{minipage}{\linewidth}
\centering
\caption*{Pressure RMSE $\times 10^{-2}$}
\scalebox{0.9}{
\begin{tabular}{@{}lcccc@{}}
\toprule
\addlinespace
\textbf{Train Dataset} & \textbf{1-step} & \textbf{50-steps} & \textbf{all-steps}& \textbf{all-steps median} \\
\addlinespace
\midrule
\addlinespace 
standard\_cylinder & $16.37 \pm 0.11$ & $10.68 \pm 0.93$ & $29.53 \pm 0.58$ & $4.14 \pm 0.2$\\
cylinder\_stretch & \bm{$11.46 \pm 0.03$} & \bm{$9.03 \pm 1.31$} & \bm{$28.65 \pm 1.87$} & $4.1 \pm 0.26$\\
cylinder\_tri\_quad & $14.38 \pm 0.15$ & $11.98 \pm 0.88$ & $30.87 \pm 2.54$ & \bm{$3.71 \pm 0.13$}\\
2cylinders & $16.46 \pm 0.37$ & $17.93 \pm 1.11$ & $35.81 \pm 0.57$ & $9.35 \pm 0.5$\\
mixed\_all & $15.7 \pm 0.18$ & $15.95 \pm 0.45$ & $30.69 \pm 0.79$ &$7.22 \pm 1.54$ \\
\textit{None} & $57.65 \pm 1.04$ & $50.74 \pm 0.73$ & $57.45 \pm 0.84$ & $22.68 \pm 1.76$\\
\addlinespace
\bottomrule
\end{tabular}}
\end{minipage}

\caption{RMSE if evaluated on \textbf{cylinder\_stretch}. \textit{None} means that a newly initialized network without any training was evaluated. The lowest RMSE of each column has been marked.}
\label{tab:rmse_stretch}

\end{table}

\begin{table}[H]
\centering

\begin{minipage}{\linewidth}
\centering
\caption*{Velocity RMSE $\times 10^{-3}$}
\scalebox{0.90}{
\begin{tabular}{@{}lcccc@{}}
\toprule
\addlinespace
\textbf{Train Dataset} & \textbf{1-step} & \textbf{50-steps} & \textbf{all-steps}& \textbf{all-steps median} \\
\addlinespace
\midrule
\addlinespace
standard\_cylinder & $7.74 \pm 0.19$ & $80.62 \pm 1.5$ & $163.55 \pm 11.32$ & $53.12 \pm 10.49$ \\
cylinder\_stretch & $5.48 \pm 0.11$ & $84.47 \pm 4.18$ & $188.36 \pm 3.39$ & $59.97 \pm 10.07$ \\
cylinder\_tri\_quad & \bm{$3.58 \pm 0.03$} & \bm{$48.09 \pm 4.94$} & \bm{$154.45 \pm 1.73$} & \bm{$51.1 \pm 5.53$}\\
2cylinders & $6.81 \pm 0.09$ & $77.47 \pm 2.13$ & $183.06 \pm 3.36$ & $62.54 \pm 10.71$\\
mixed\_all & $3.76 \pm 0.06$ & $59.59 \pm 7.42$ & $197.34 \pm 3.37$ & $63.13 \pm 2.65$ \\
\textit{None} & $45.22 \pm 1.03$ & $624.81 \pm 216.15$ & $2032.21 \pm 1206.92$ & $2024.38 \pm 1217.24$ \\
\addlinespace
\bottomrule
\end{tabular}}
\end{minipage}

\begin{minipage}{\linewidth}
\centering
\caption*{Pressure RMSE $\times 10^{-2}$}
\scalebox{0.90}{
\begin{tabular}{@{}lcccc@{}}
\toprule
\addlinespace
\textbf{Train Dataset} & \textbf{1-step} & \textbf{50-steps} & \textbf{all-steps}& \textbf{all-steps median} \\
\addlinespace
\midrule
\addlinespace
standard\_cylinder & $12.24 \pm 0.07$ & $13.8 \pm 0.06$ & $20.95 \pm 0.33$ & $5.22 \pm 0.37$ \\
cylinder\_stretch & $12.48 \pm 0.45$ & $14.29 \pm 0.17$ & $23.7 \pm 0.39$ & $6.59 \pm 1.5$ \\
cylinder\_tri\_quad & \bm{$9.62 \pm 0.2$} & \bm{$9.96 \pm 0.49$} & \bm{$18.79 \pm 0.7$} & \bm{$5.09 \pm 0.45$} \\
2cylinders & $16.35 \pm 0.07$ & $19.1 \pm 0.29$ & $24.59 \pm 0.18$ & $10.21 \pm 0.65$ \\
mixed\_all & $14.38 \pm 0.26$ & $15.44 \pm 1.14$ & $24.04 \pm 0.22$ & $8.2 \pm 0.47$ \\
\textit{None} & $48.06 \pm 2.78$ & $46.23 \pm 1.45$ & $50.79 \pm 0.7$ & $37.96 \pm 0.87$\\
\addlinespace
\bottomrule
\end{tabular}}
\end{minipage}

\caption{RMSE if evaluated on \textbf{cylinder\_tri\_quad}. \textit{None} means that a newly initialized network without any training was evaluated. The lowest RMSE of each column has been marked.}
\label{tab:rmse_quad}

\end{table}

\begin{table}[H]
\centering

\begin{minipage}{\linewidth}
\centering
\caption*{Velocity RMSE $\times 10^{-3}$}
\scalebox{0.90}{
\begin{tabular}{@{}lcccc@{}}
\toprule
\addlinespace
\textbf{Train Dataset} & \textbf{1-step} & \textbf{50-steps} & \textbf{all-steps}& \textbf{all-steps median} \\
\addlinespace
\midrule
\addlinespace
standard\_cylinder & $7.79 \pm 0.2$ & $54.37 \pm 2.64$ & $191.26 \pm 10.76$ & $101.61 \pm 11.7$ \\
cylinder\_stretch & $7.31 \pm 0.06$ & $62.11 \pm 2.79$ & $191.01 \pm 8.7$ & $110.08 \pm 17.34$  \\
cylinder\_tri\_quad & $7.01 \pm 0.11$ & $61.59 \pm 1.03$ & $214.92 \pm 13.39$ & $121.21 \pm 13.65$ \\
2cylinders & \bm{$3.8 \pm 0.03$} & \bm{$49.92 \pm 2.53$} & \bm{$172.97 \pm 22.63$}  & \bm{$86.5 \pm 6.04$}\\
mixed\_all & $4.64 \pm 0.02$ & $65.34 \pm 6.9$ & $198.22 \pm 15.02$ & $103.37 \pm 10.99$ \\
\textit{None} & $47.8 \pm 2.06$ & $878.86 \pm 177.63$ & $2861.77 \pm 457.12$ & $2864.81 \pm 472.92$ \\
\addlinespace
\bottomrule
\end{tabular}}
\end{minipage}

\begin{minipage}{\linewidth}
\centering
\caption*{Pressure RMSE $\times 10^{-2}$}
\scalebox{0.90}{
\begin{tabular}{@{}lcccc@{}}
\toprule
\addlinespace
\textbf{Train Dataset} & \textbf{1-step} & \textbf{50-steps} & \textbf{all-steps}& \textbf{all-steps median} \\
\addlinespace
\midrule
\addlinespace
standard\_cylinder & $25.68 \pm 0.36$ & $18.73 \pm 0.12$ & $31.57 \pm 0.78$ & $15.54 \pm 0.39$ \\
cylinder\_stretch & $24.8 \pm 0.02$ & \bm{$18.6 \pm 0.09$} & $31.12 \pm 0.94$&  $16.77 \pm 1.3$ \\
cylinder\_tri\_quad & $25.46 \pm 0.29$ & $19.85 \pm 0.16$ & $33.8 \pm 1.34$ & $18.3 \pm 1.42$ \\
2cylinders & \bm{$22.43 \pm 0.25$} & $20.43 \pm 0.31$ & \bm{$29.63 \pm 1.86$}  & \bm{$15.51 \pm 1.18$}\\
mixed\_all & $23.71 \pm 0.33$ & $21.93 \pm 0.57$ & $31.96 \pm 1.52$ & $18.4 \pm 0.76$\\
\textit{None} & $57.04 \pm 0.78$ & $53.15 \pm 2.59$ & $58.17 \pm 1.86$ & $39.02 \pm 1.33$ \\
\addlinespace
\bottomrule
\end{tabular}}
\end{minipage}

\caption{RMSE if evaluated on \textbf{2cylinders}. \textit{None} means that a newly initialized network without any training was evaluated. The lowest RMSE of each column has been marked.}
\label{tab:rmse_2}

\end{table}

\begin{table}[H]
\centering

\begin{minipage}{\linewidth}
\centering
\caption*{Velocity RMSE $\times 10^{-3}$}
\scalebox{0.90}{
\begin{tabular}{@{}lcccc@{}}
\toprule
\addlinespace
\textbf{Train Dataset} & \textbf{1-step} & \textbf{50-steps} & \textbf{all-steps}& \textbf{all-steps median} \\
\addlinespace
\midrule
\addlinespace
standard\_cylinder & $18.99 \pm 0.04$ & $172.27 \pm 1.92$ & $305.93 \pm 6.97$& $133.08 \pm 7.23$ \\
cylinder\_stretch & $13.0 \pm 0.06$ & $175.75 \pm 3.76$ & $310.26 \pm 1.72$ & $150.07 \pm 18.13$\\
cylinder\_tri\_quad & $9.23 \pm 0.12$ & $132.15 \pm 5.87$ & $320.14 \pm 16.5$ & \bm{$82.41 \pm 17.75$}\\
2cylinders & $16.27 \pm 0.23$ & $184.82 \pm 3.45$ & $317.82 \pm 11.12$ & $141.39 \pm 16.34$ \\
mixed\_all & \bm{$7.39 \pm 0.1$} & \bm{$113.16 \pm 3.18$} & \bm{$299.35 \pm 14.67$} & $92.7 \pm 2.04$\\
\textit{None} & $60.08 \pm 2.74$ & $956.38 \pm 392.02$ & $3394.79 \pm 1824.51$ & $3439.83 \pm 1862.33$ \\
\addlinespace
\bottomrule
\end{tabular}}
\end{minipage}

\begin{minipage}{\linewidth}
\centering
\caption*{Pressure RMSE $\times 10^{-2}$}
\scalebox{0.90}{
\begin{tabular}{@{}lcccc@{}}
\toprule
\addlinespace
\textbf{Train Dataset} & \textbf{1-step} & \textbf{50-steps} & \textbf{all-steps}& \textbf{all-steps median} \\
\addlinespace
\midrule
\addlinespace 
standard\_cylinder & $52.97 \pm 0.7$ & $52.41 \pm 0.77$ & $68.22 \pm 0.94$ & $11.47 \pm 0.61$\\
cylinder\_stretch & $40.59 \pm 0.22$ & $47.55 \pm 0.62$ & $64.19 \pm 0.46$ & $13.83 \pm 3.62$\\
cylinder\_tri\_quad & $36.37 \pm 0.7$ & $39.12 \pm 1.73$ & $61.56 \pm 1.79$ & \bm{$11.43 \pm 3.33$}\\
2cylinders & $44.38 \pm 0.47$ & $54.2 \pm 1.05$ & $68.97 \pm 0.51$  & $15.95 \pm 1.43$\\
mixed\_all & \bm{$32.07 \pm 0.39$} & \bm{$36.96 \pm 0.79$} & \bm{$59.23 \pm 2.07$}& $14.73 \pm 2.27$ \\
\textit{None} & $90.13 \pm 1.46$ & $80.7 \pm 1.71$ & $90.04 \pm 1.31$ & $32.46 \pm 4.51$ \\
\addlinespace
\bottomrule
\end{tabular}}
\end{minipage}

\caption{RMSE if evaluated on \textbf{mixed\_all}. \textit{None} means that a newly initialized network without any training was evaluated. The lowest RMSE of each column has been marked.}
\label{tab:rmse_mixed}

\end{table}

\subsubsection{General observations}
Before we draw any conclusions, we will put into context what these RMSE values mean. All models produce physically reasonable predictions on every dataset and the following observations apply to all training-evaluation dataset pairs. There are three main ways in which the MGN prediction differs from our FEM simulation. 

\begin{figure}[H]
  \centering
  \includegraphics[width=0.65\textwidth]{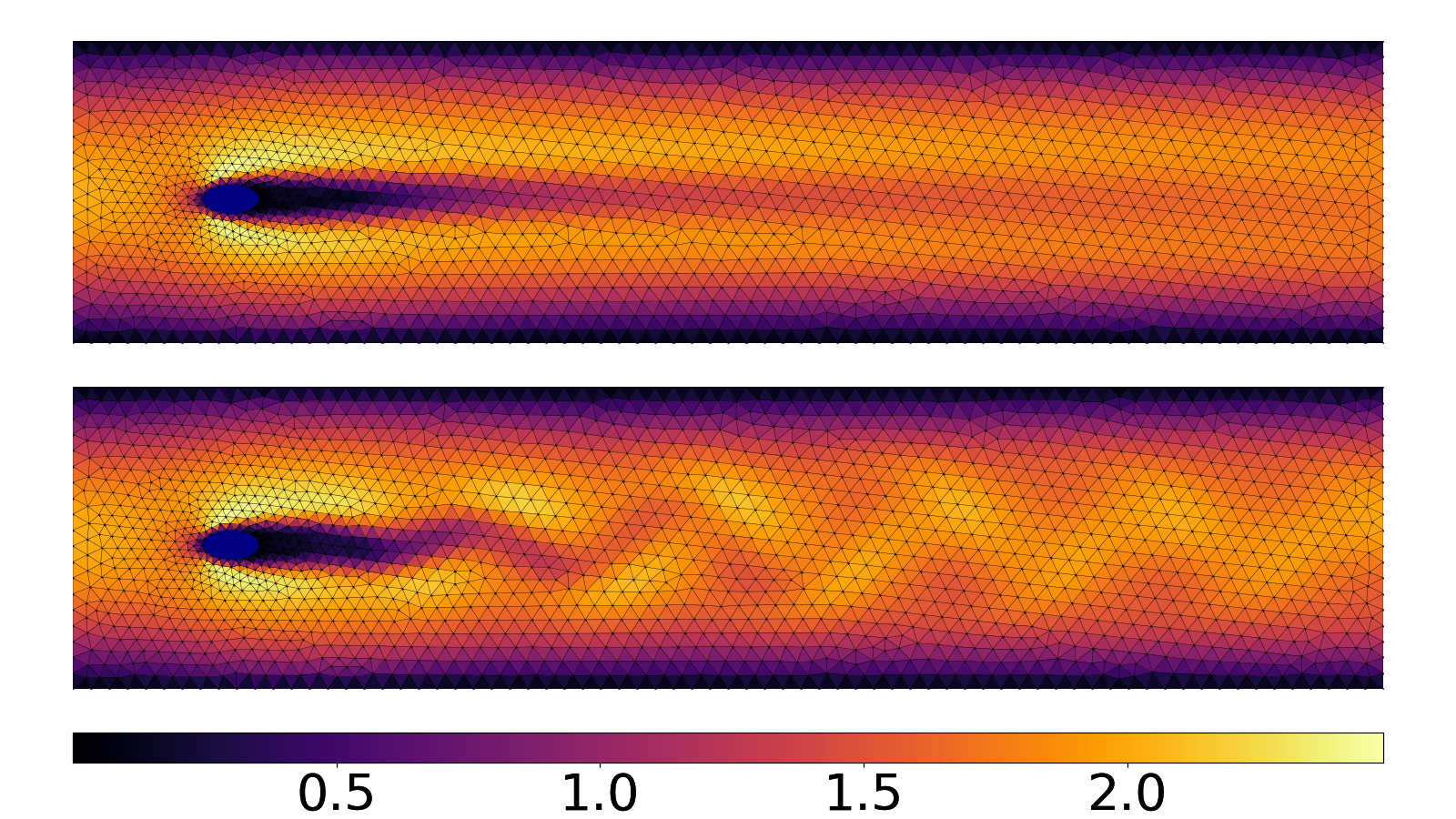}
  \caption{Plot of the Euclidean norm of the velocity field at the final time step for a FEM simulation (bottom) and the corresponding MGN prediction (top) of a mesh coming from the \textbf{cylinder\_stretch} dataset. The prediction has an all-steps RMSE of circa $0.076$ and stems from an MGN that was trained on \textbf{cylinder\_stretch}.}
  \label{fig:no_vortex_shedding}
\end{figure}

Firstly, the MGN prediction for all models sometimes does not predict vortex shedding. This is more likely to happen for inflow peaks that are at the border where vortex shedding starts to occur. Fig. \ref{fig:no_vortex_shedding} shows an example of this. All figures presented in this and the following sections show only the pressure or velocity field prediction at the final time step. However, for each figure, we uploaded the full simulation animation as a gif file to this work's GitHub repository \cite{github_implementation}. 

\begin{figure}[H]
\subfloat[][]
{\includegraphics[width=0.45\textwidth]{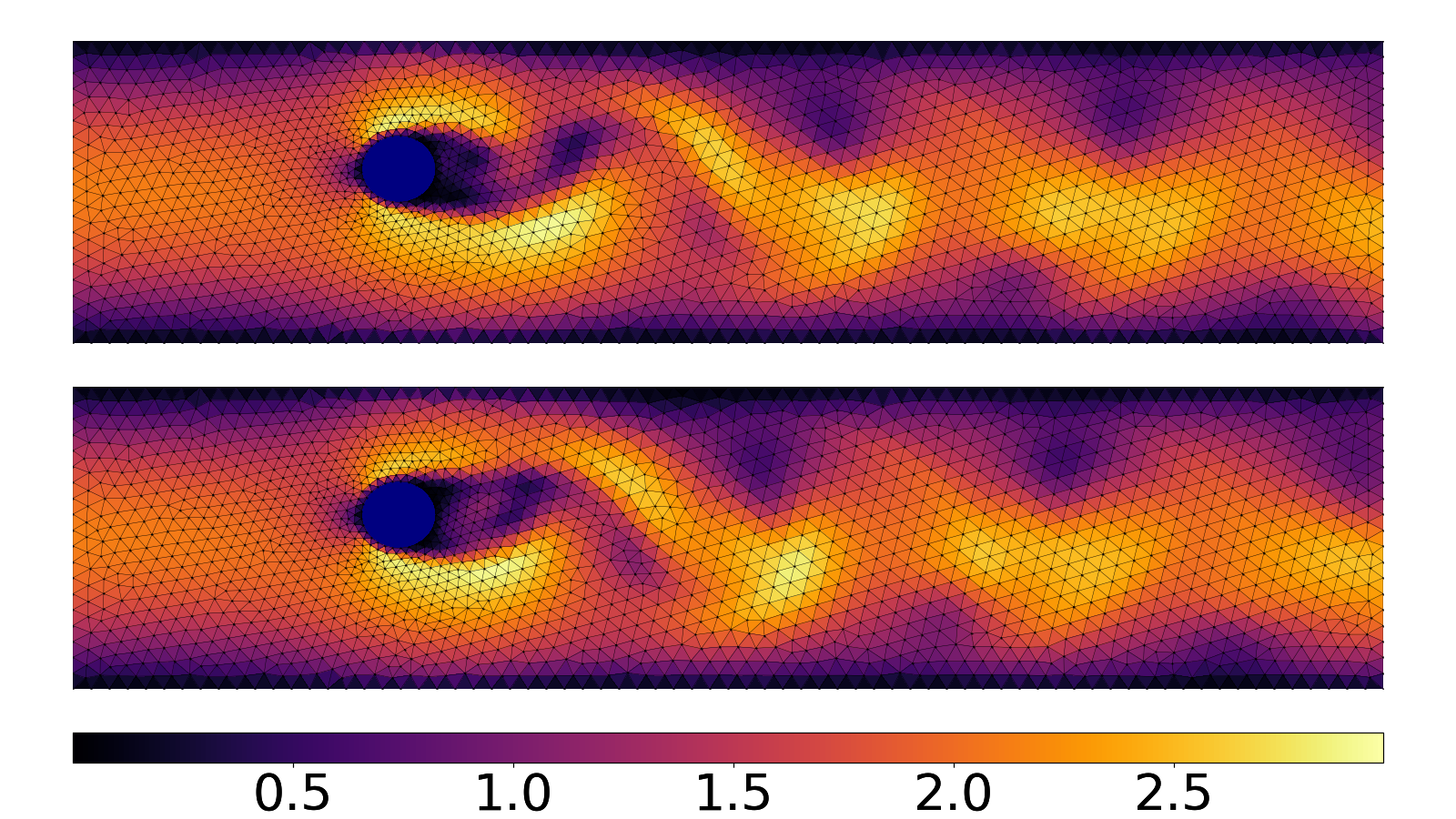}}
\hfill
\subfloat[][]
{\includegraphics[width=0.45\textwidth]{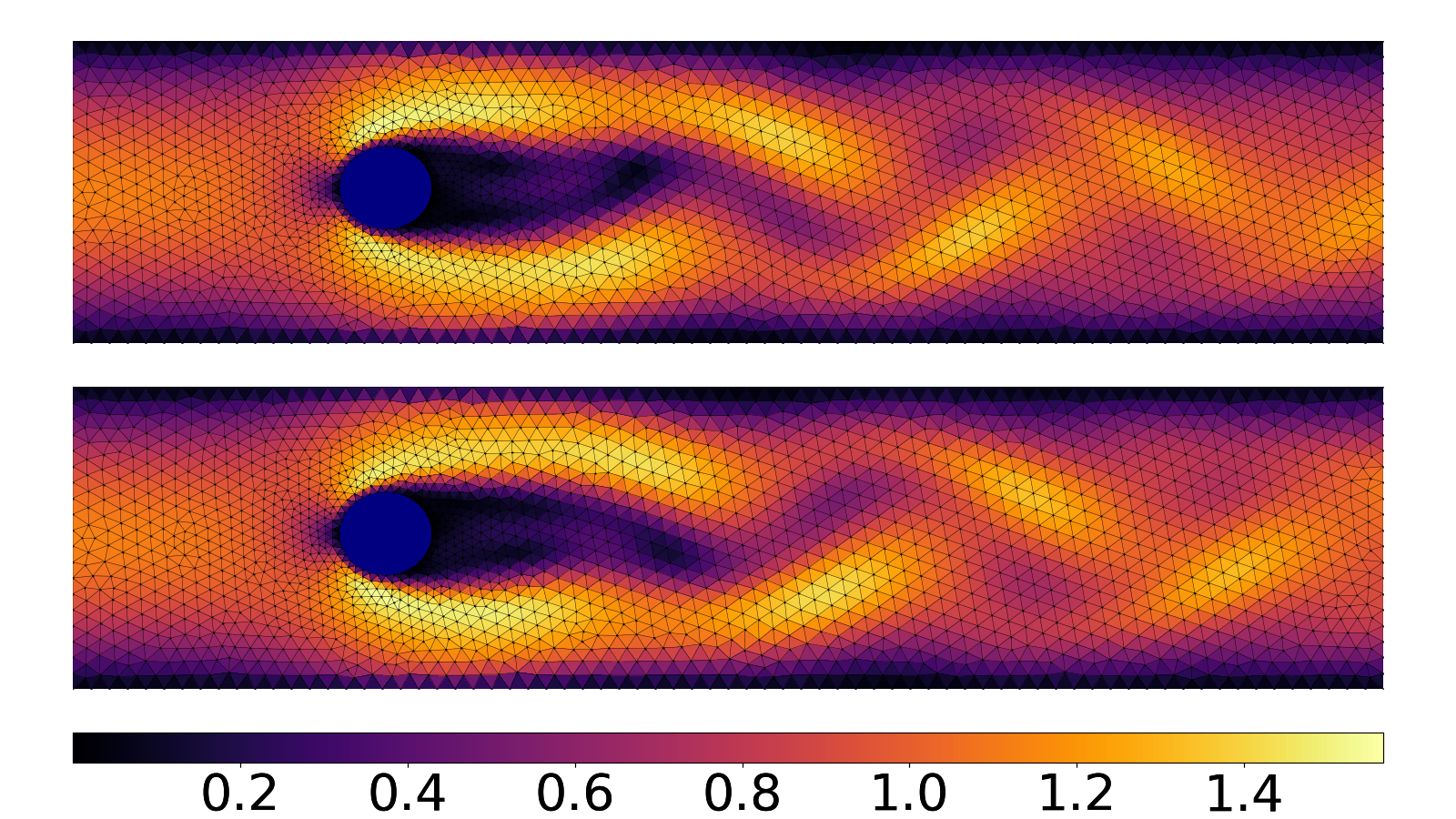}}
\caption{Plot of the Euclidean norm of the velocity field at the final time step for two FEM simulations (bottom) and their corresponding MGN predictions (top) of two meshes coming from the \textbf{standard\_cylinder} dataset. The prediction (a) has an all-steps RMSE of circa $0.193$ and (b) has an all-steps RMSE of circa $0.086$. Both predictions stem from an MGN that was trained on \textbf{standard\_cylinder}.}
\label{fig:out_of_sync_vortex}
\end{figure}

Secondly, the MGN sometimes fails to estimate the propagation speed of the vortices. This causes the MGN prediction to look out of sync at later time steps. This is especially prevalent for high inflow peaks. Fig. \ref{fig:out_of_sync_vortex} shows two examples of this. 
Fig. \ref{fig:out_of_sync_vortex} also demonstrates that the RMSE does not always capture how qualitatively wrong the prediction is as a higher inflow peak causes the average magnitude of the velocity field and thereby its error to be higher.

\begin{figure}[H]
  \centering
  \includegraphics[width=0.65\textwidth]{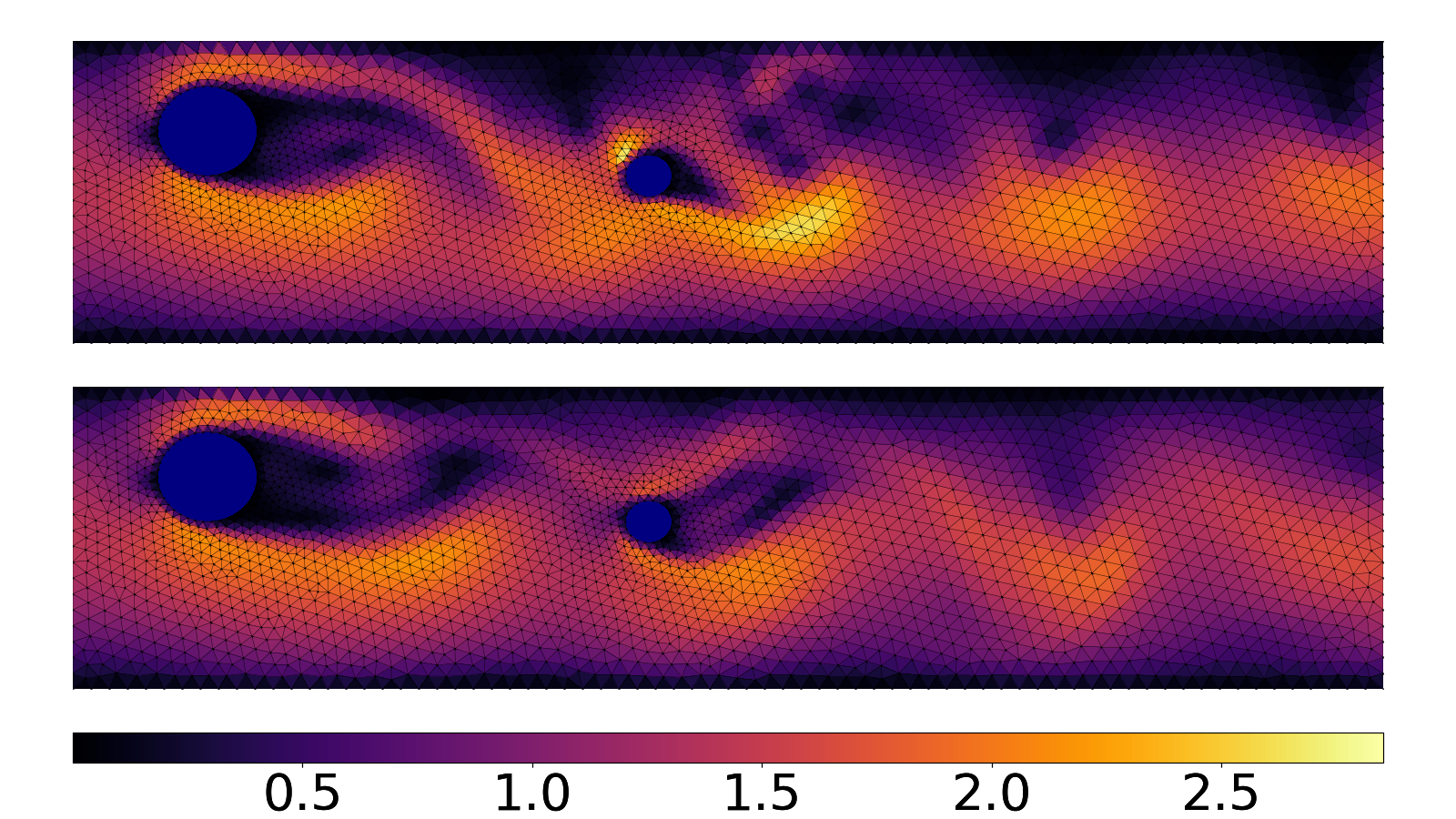}
  \caption{Plot of the Euclidean norm of the velocity field at the final time step for a FEM simulation (bottom) and the corresponding MGN prediction (top) of a mesh coming from the \textbf{2cylinders} dataset. The prediction has an all-steps RMSE of circa $0.318$ and stems from an MGN that was trained on \textbf{standard\_cylinder}. Note that the vortex pattern at the smaller circle differs from the IPCS solution as the MGN predicts pairs of vortices whereas the IPCS does not.}
  \label{fig:wrong_vortex_pattern}
\end{figure}

Thirdly, when the MGN (and this is mostly the case for unseen shapes) predicts vortex shedding, it predicts a wrong pattern of the vortices. Fig. \ref{fig:wrong_vortex_pattern} shows an example of this.

\begin{figure}[H]
  \centering
  \includegraphics[width=0.65\textwidth]{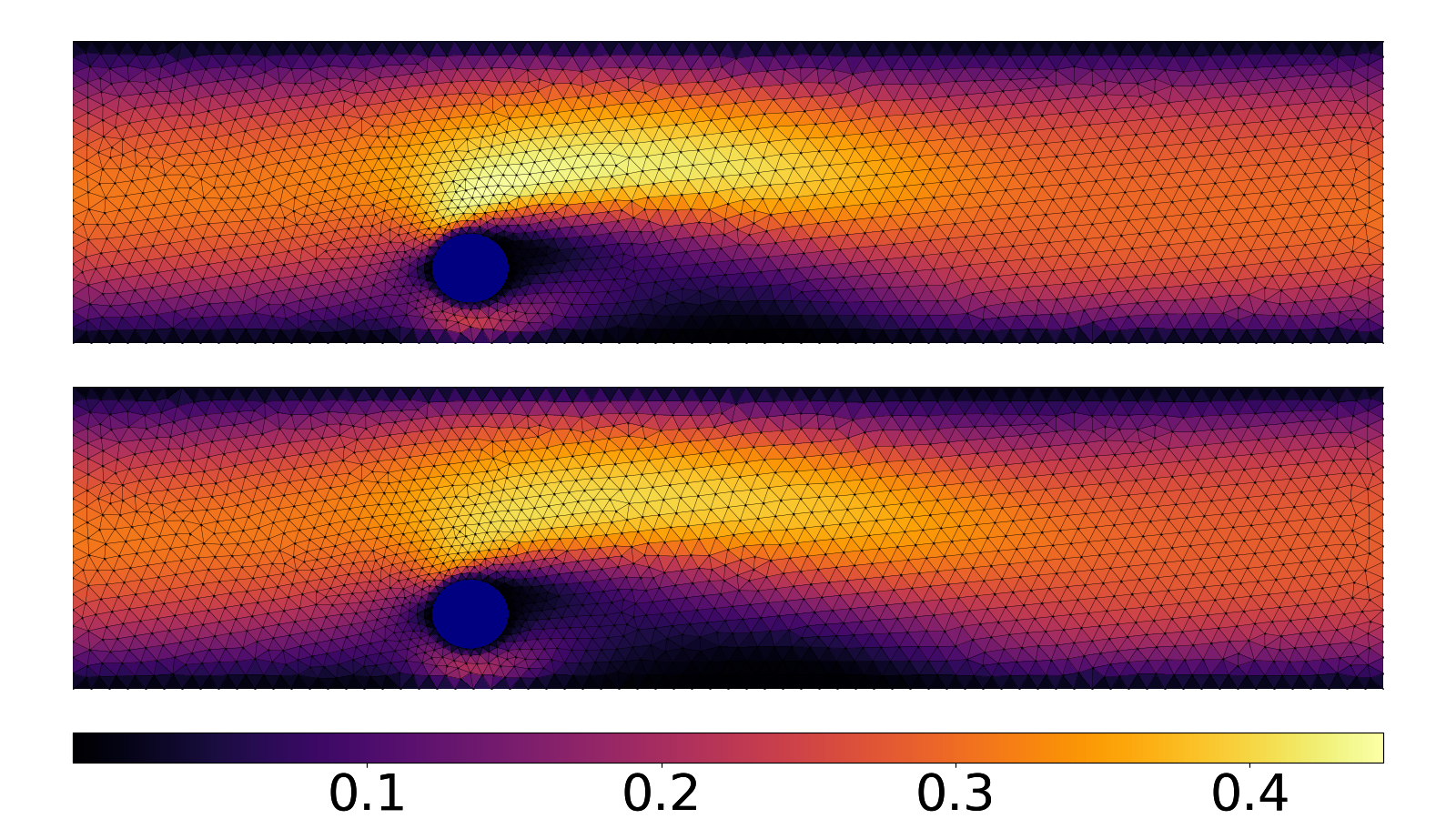}
  \caption{Plot of the Euclidean norm of the velocity field at the final time step for a FEM simulation (bottom) and the corresponding MGN prediction (top) of a mesh coming from the \textbf{standard\_cylinder} dataset. The prediction has an all-steps RMSE of circa $0.012$ and stems from an MGN that was trained on \textbf{standard\_cylinder}.}
  \label{fig:low_inflow_regime}
\end{figure}

Unsurprisingly, all models thrive in handling the low inflow peak regime where no vortex shedding occurs. Fig. \ref{fig:low_inflow_regime} shows an example of this.

\begin{figure}[H]
\subfloat[][Velocity]
{\includegraphics[width=0.45\textwidth]{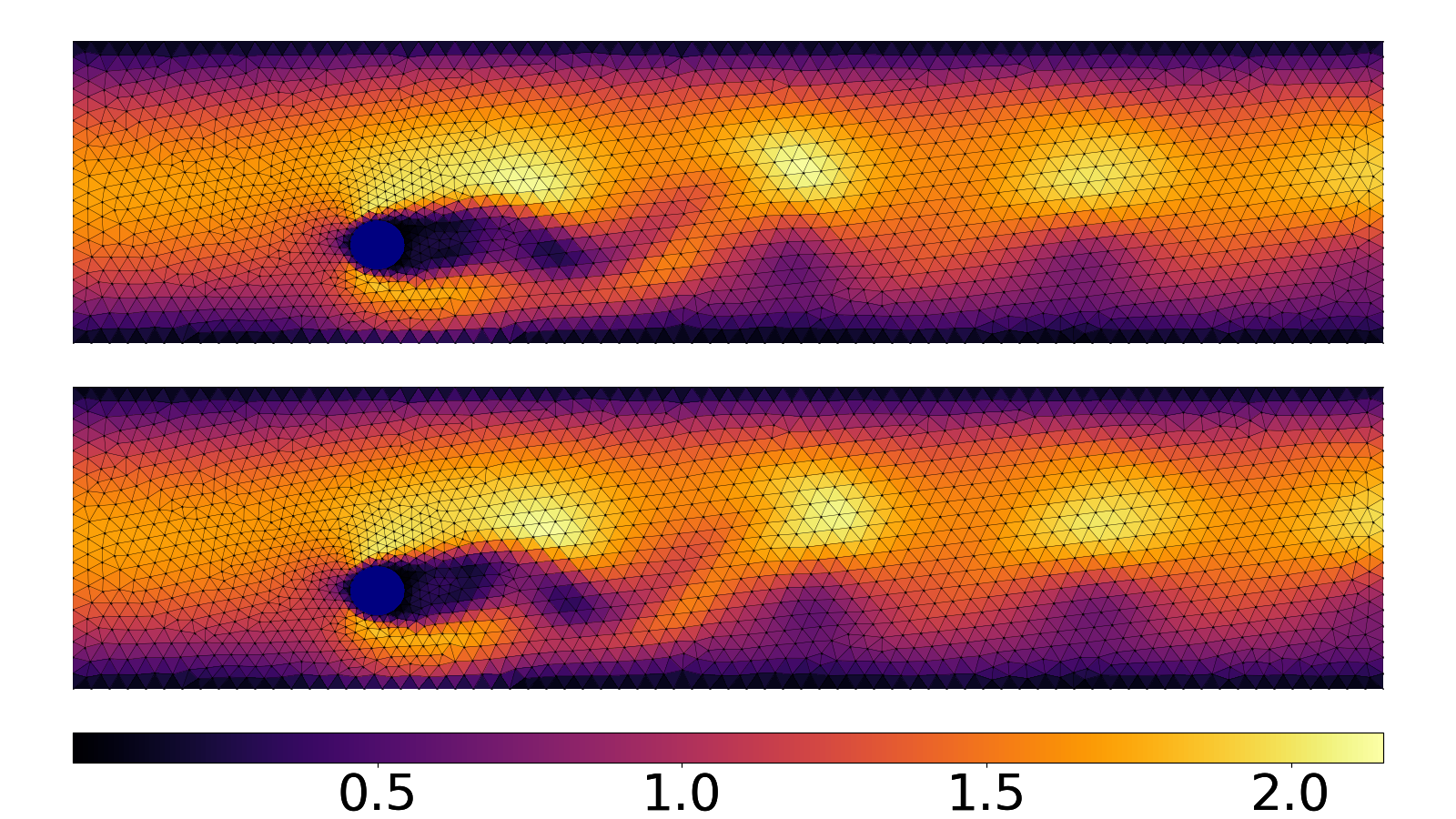}}
\hfill
\subfloat[][Pressure]
{\includegraphics[width=0.45\textwidth]{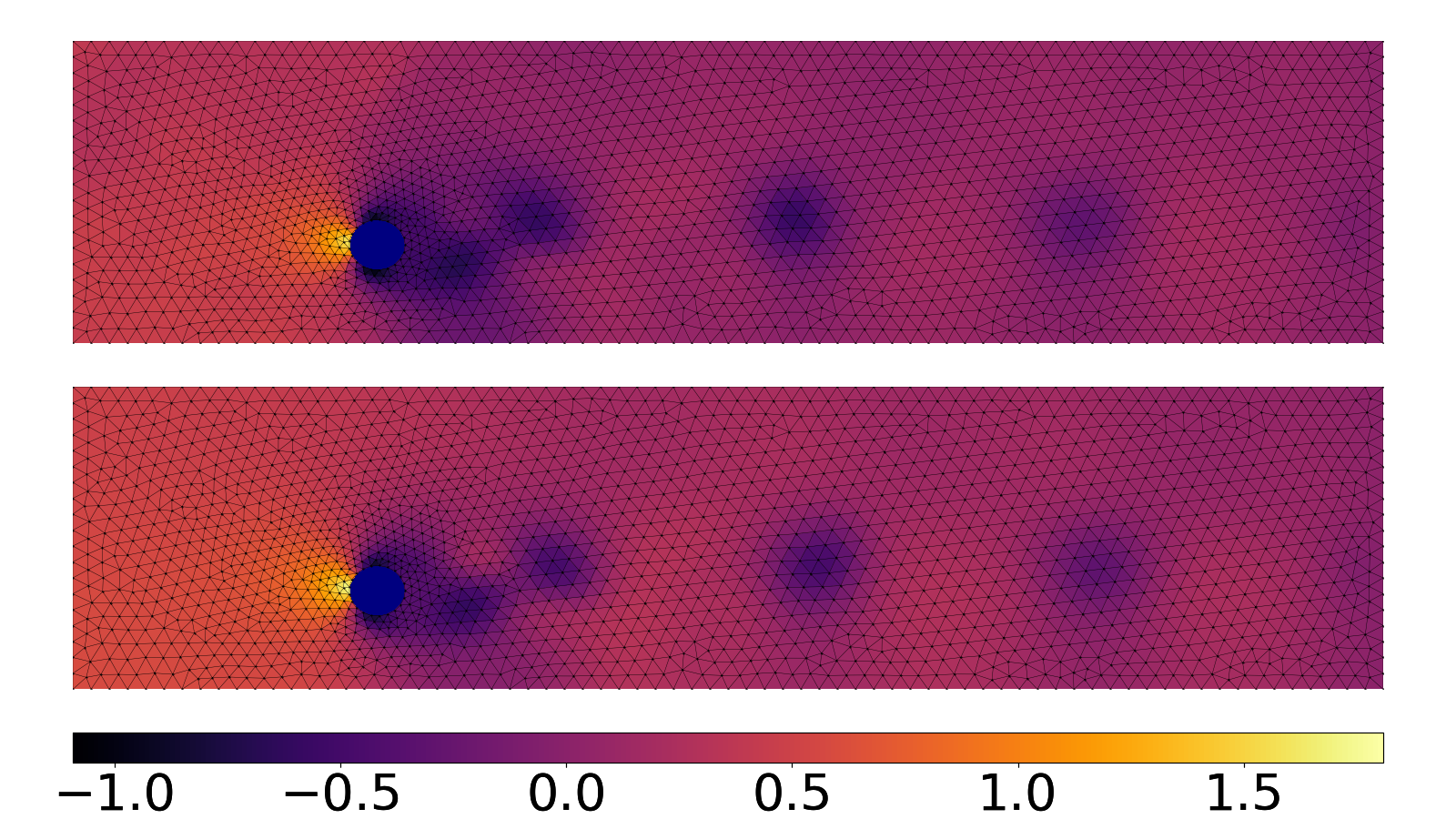}}
\caption{Plot of the pressure field (right) and the Euclidean norm of the velocity field (left) at the final time step for a FEM simulation (bottom) and its corresponding MGN prediction (top) of one mesh coming from the \textbf{standard\_cylinder} dataset. The velocity prediction (a) has an all-steps RMSE of roughly $0.032$ and the pressure prediction (b) has an all-steps RMSE of approximately $0.016$. Both predictions stem from an MGN that was trained on \textbf{standard\_cylinder}.}
\label{fig:pressure_standard_cyl}
\end{figure}

Thus far we have only shown the velocity field and not the pressure field. This is because the pressure field is uninteresting as its deviation from the FEM pressure solution directly correlates with the deviation from the velocity field since the MGN directly predicts the pressure from the velocity field. Fig. \ref{fig:pressure_standard_cyl} shows the pressure field together with the velocity field and their prediction for an average \textbf{cylinder\_flow} instance.

\subsubsection{Predictions for coinciding train and test datasets}
\begin{figure}[H]
\subfloat[][Dataset: \textbf{standard\_cylinder}, RMSE: $0.029$]
{\includegraphics[width=0.47\textwidth]{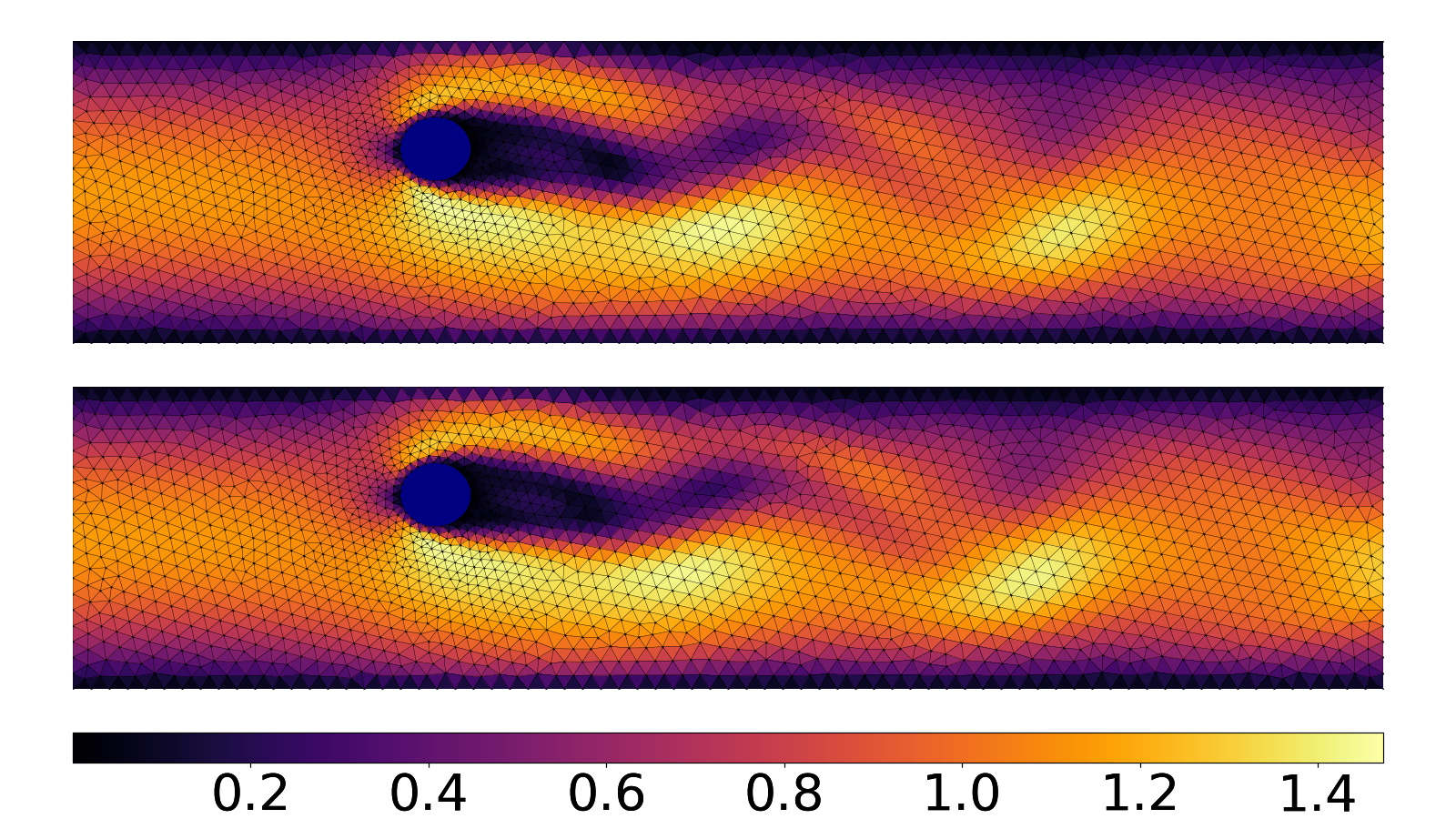}}
\hfill
\subfloat[][Dataset: \textbf{cylinder\_stretch}, RMSE: $0.075$]
{\includegraphics[width=0.47\textwidth]{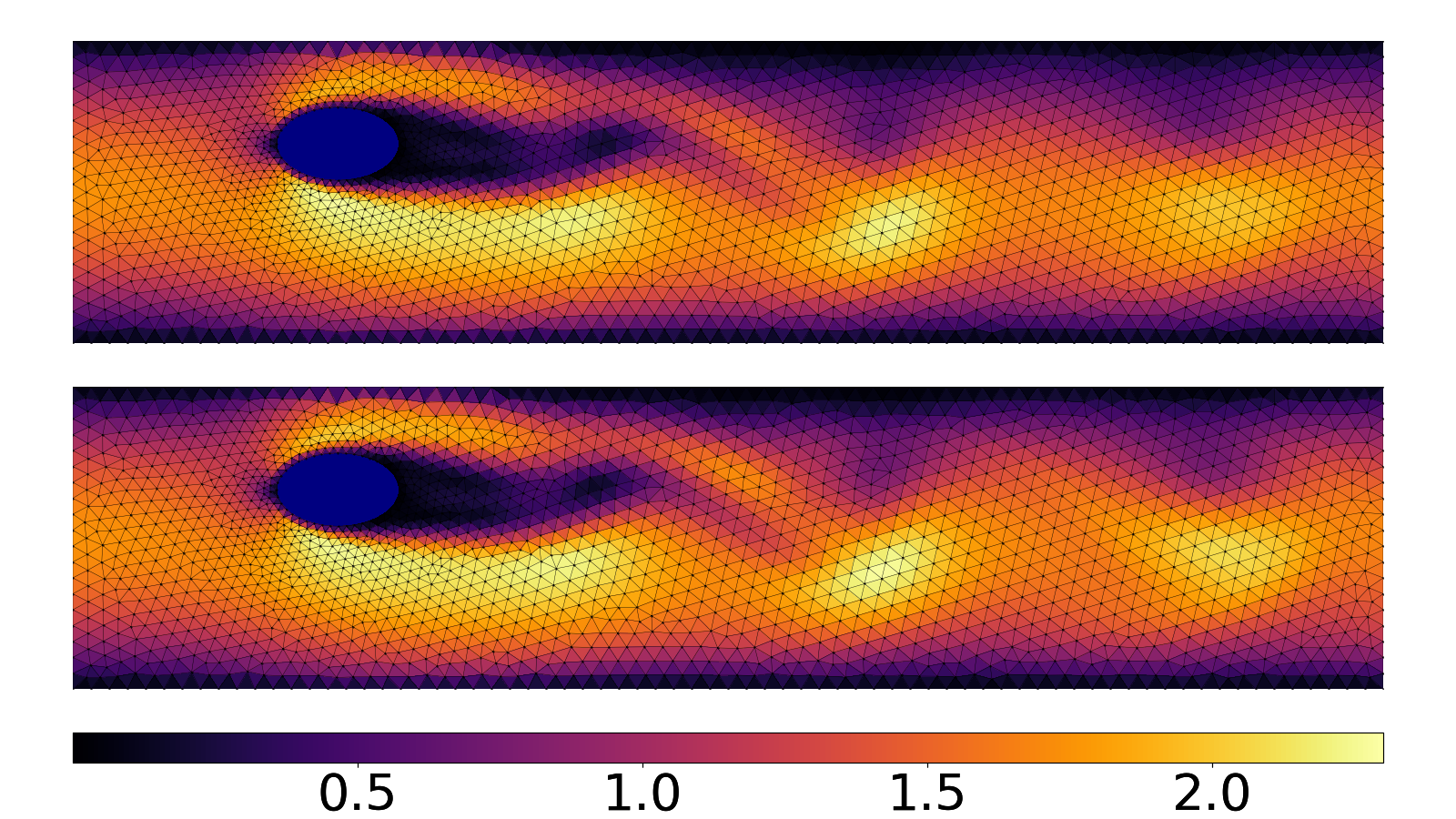}}
\hfill
\subfloat[][Dataset: \textbf{cylinder\_tri\_quad}, RMSE: $0.056$]
{\includegraphics[width=0.47\textwidth]{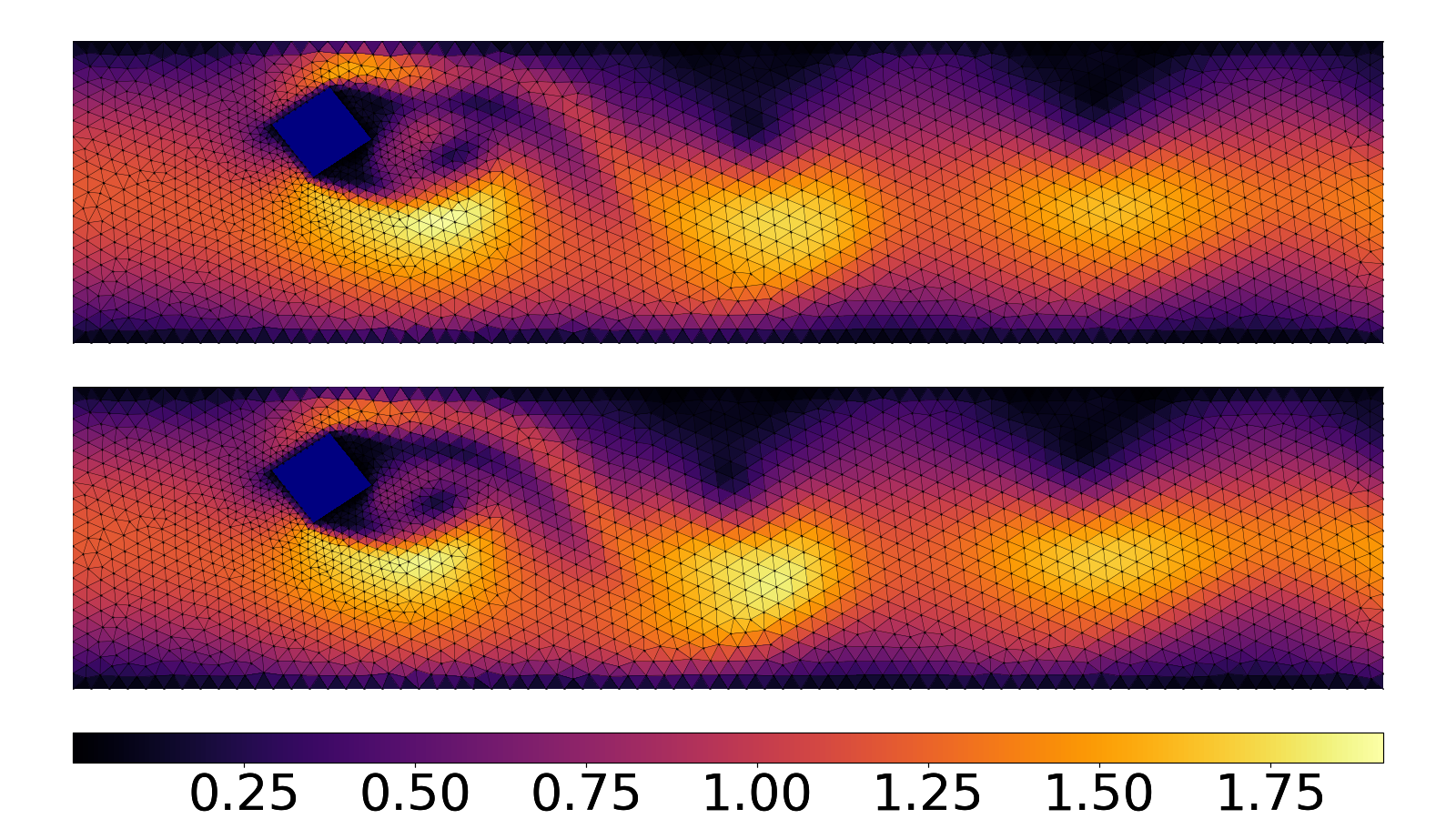}}
\hfill
\subfloat[][Dataset: \textbf{cylinder\_tri\_quad}, RMSE: $0.060$]
{\includegraphics[width=0.47\textwidth]{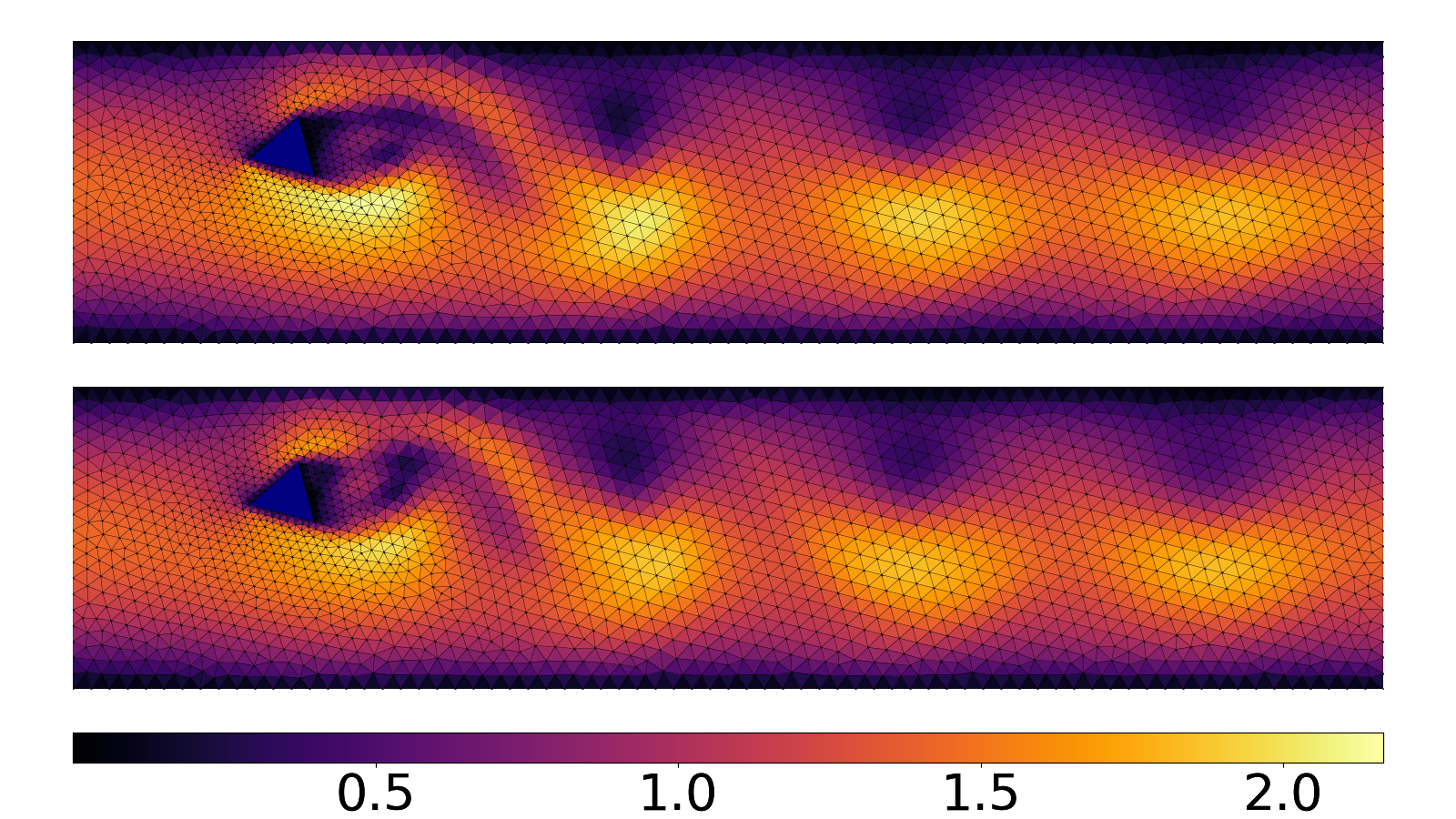}}
\hfill
\subfloat[][Dataset: \textbf{2cylinders}, RMSE: $0.111$]
{\includegraphics[width=0.47\textwidth]{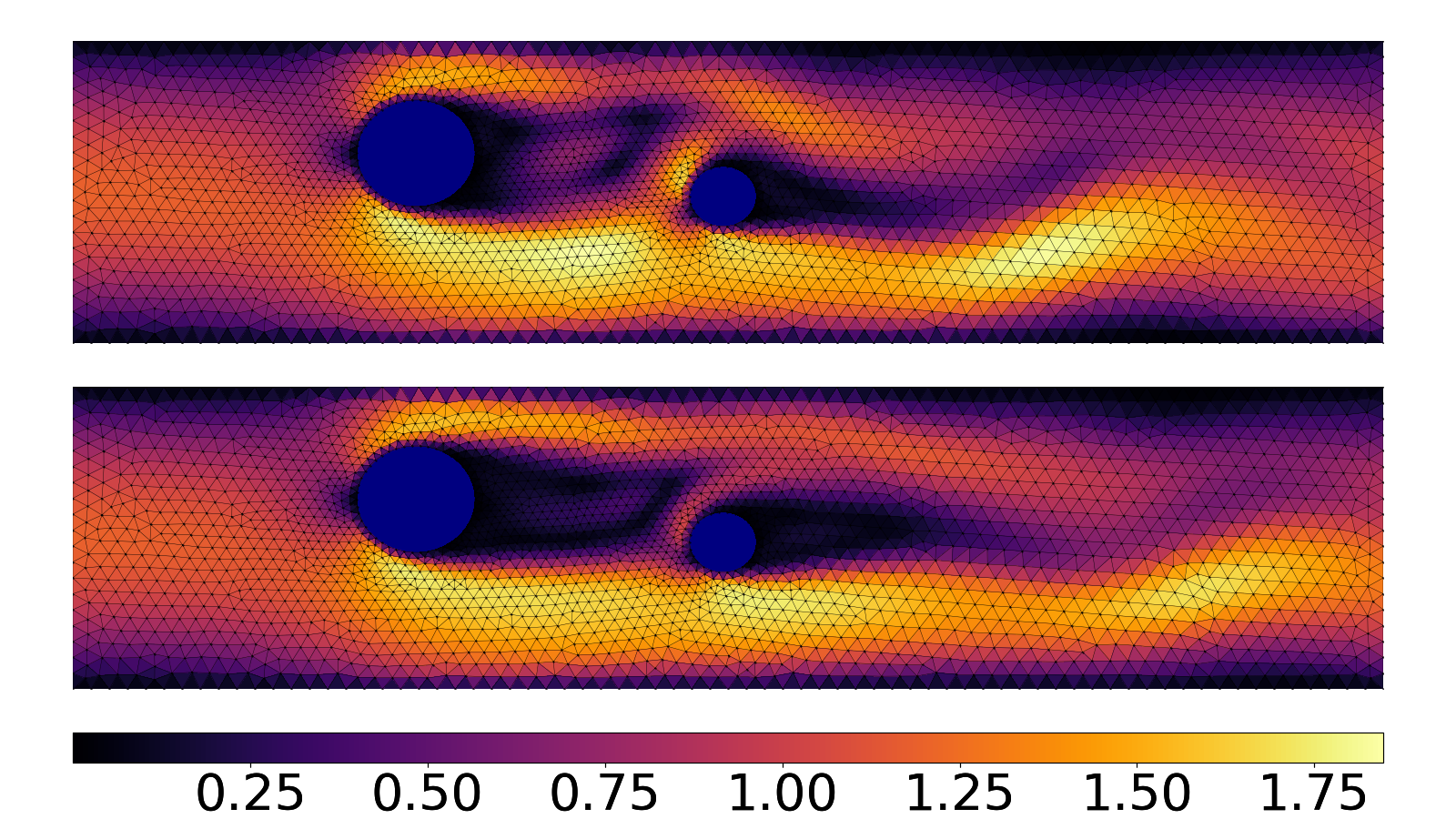}}
\hfill
\subfloat[][Dataset: \textbf{mixed\_all}, RMSE: $0.061$]
{\includegraphics[width=0.47\textwidth]{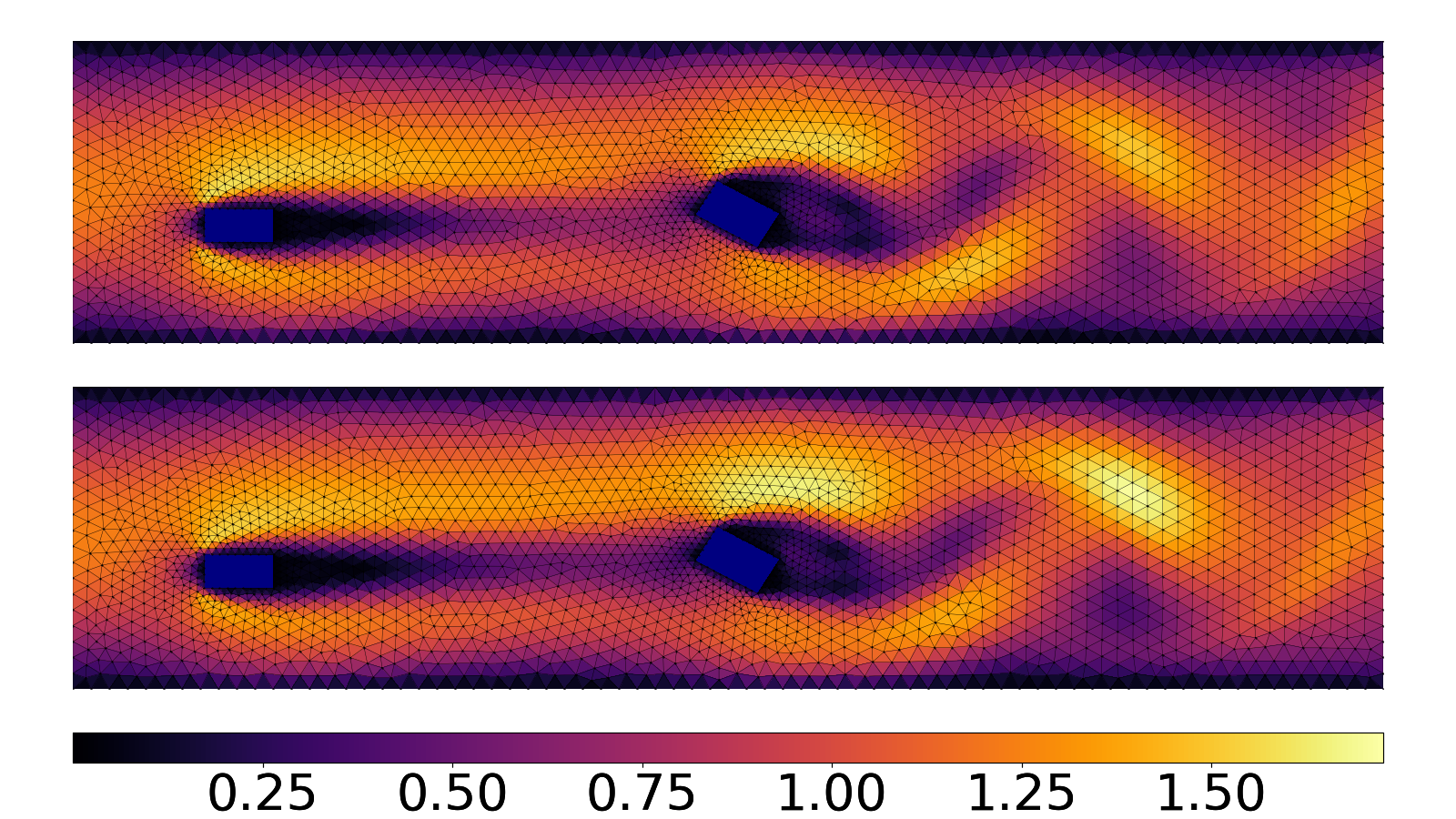}}
\caption{Plot of the Euclidean norm of the velocity field at the final time step for six FEM simulations (bottom) and their corresponding MGN predictions (top) of six meshes coming from the dataset mentioned in the corresponding subcaption. Each subcaption also contains the all-steps RMSE for the corresponding simulation rounded to three decimal places. The predictions stem from MGNs that were trained on the same dataset on which the predictions were performed.
}
\label{fig:same_set_sims}
\end{figure}

We observed that if the MGN is trained on the same dataset it is evaluated on and it predicts vortex shedding, then these vortices have mostly the same characteristics as the IPCS solution which is only partially the case for differing train and evaluation datasets as we shall see later. Fig. \ref{fig:same_set_sims} shows one simulation and its prediction for each of the five datasets when the network was trained on the same dataset. The simulations were picked to have an RMSE near the RMSE median for that dataset.
 
\subsubsection{Predictions for differing train and test datasets}
\begin{figure}[H]
\subfloat[][Dataset: \textbf{cylinder\_tri\_quad}, RMSE: $0.118$. Vortex pattern \textbf{correctly} predicted.]
{\includegraphics[width=0.47\textwidth]{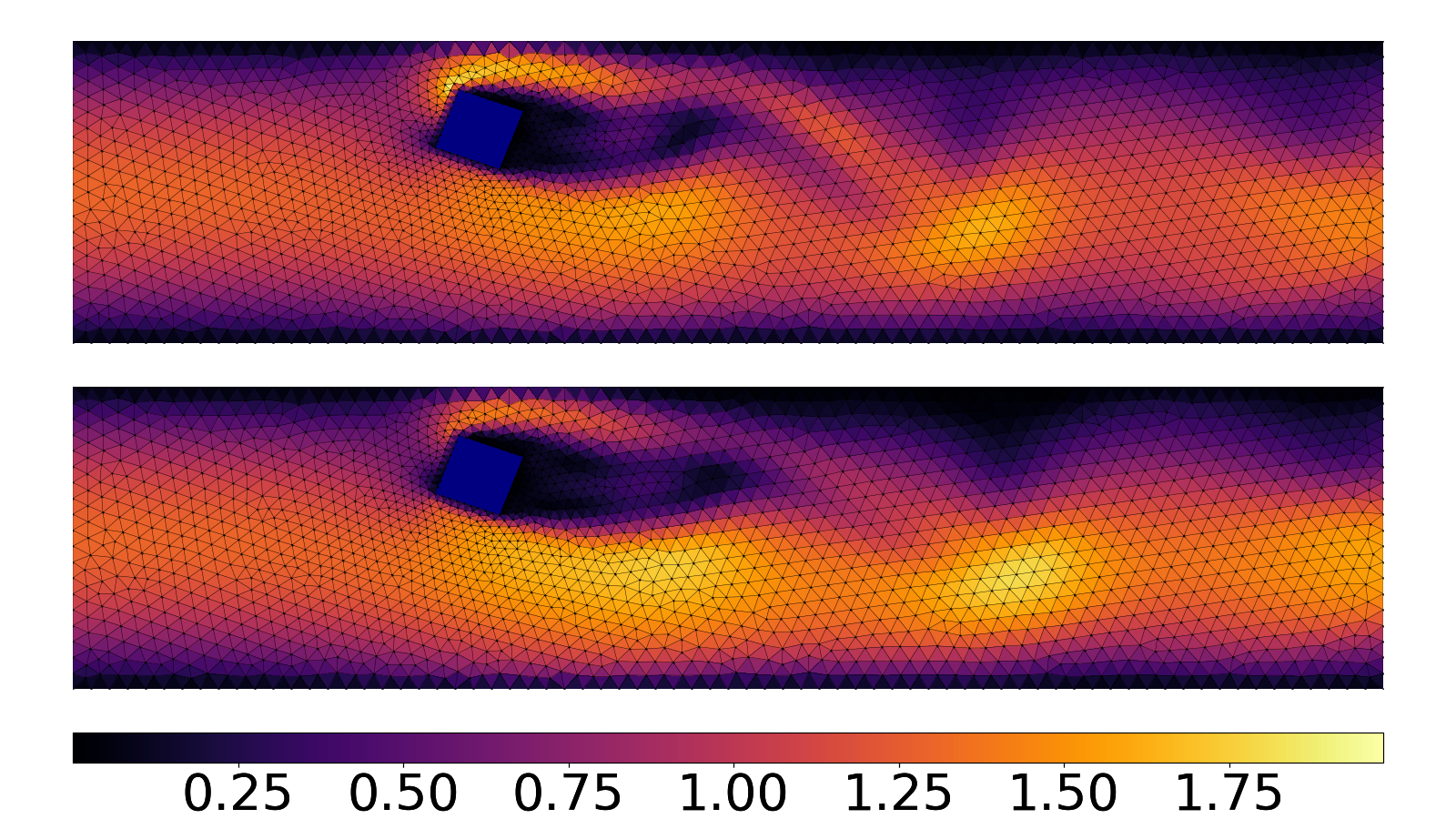}}
\hfill
\subfloat[][Dataset: \textbf{cylinder\_tri\_quad}, RMSE: $0.196$. Vortex pattern \textbf{falsely} predicted.]
{\includegraphics[width=0.47\textwidth]{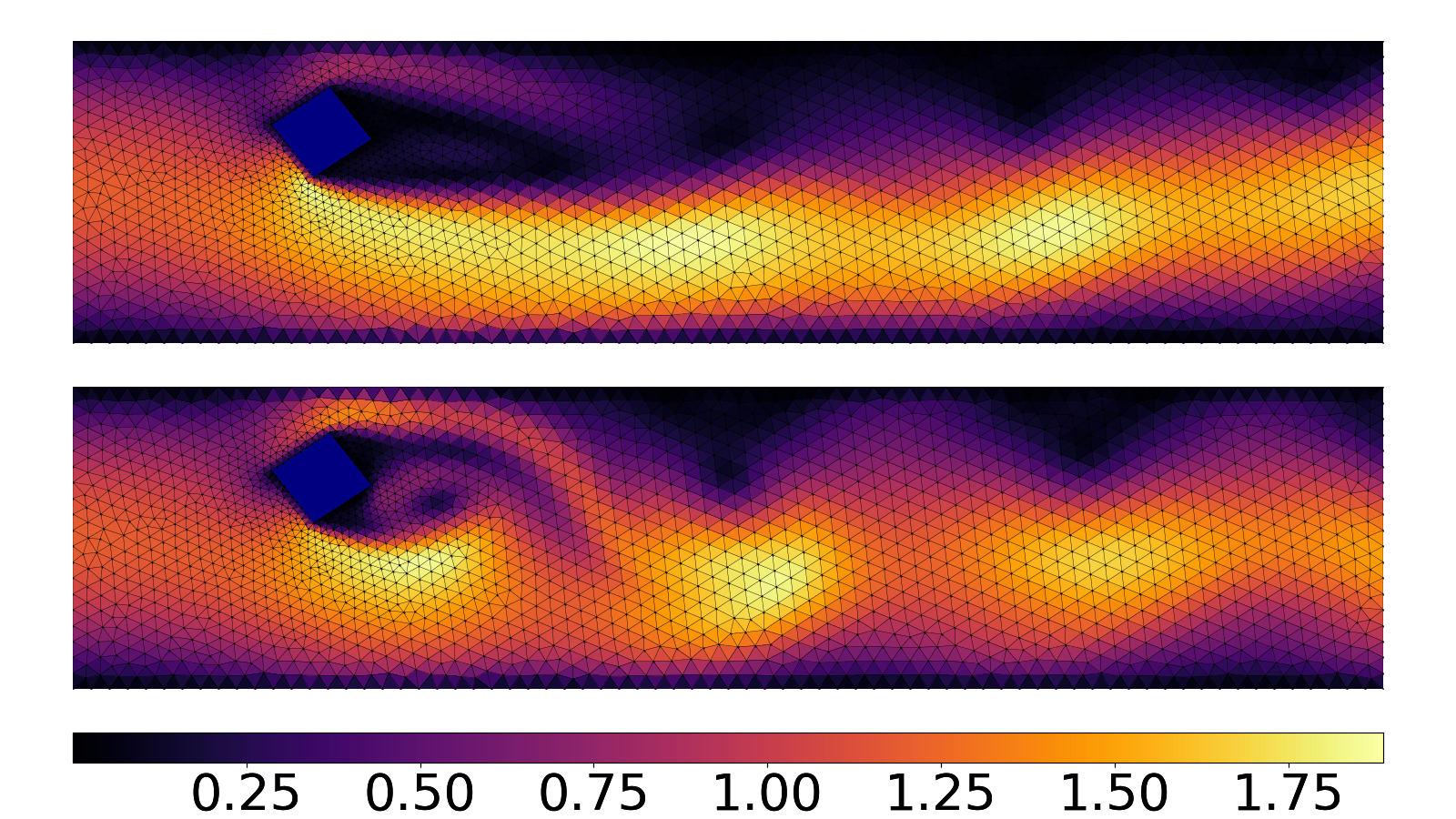}}
\hfill
\subfloat[][Dataset: \textbf{cylinder\_tri\_quad}, RMSE: $0.197$. Vortex pattern \textbf{correctly} predicted.]
{\includegraphics[width=0.47\textwidth]{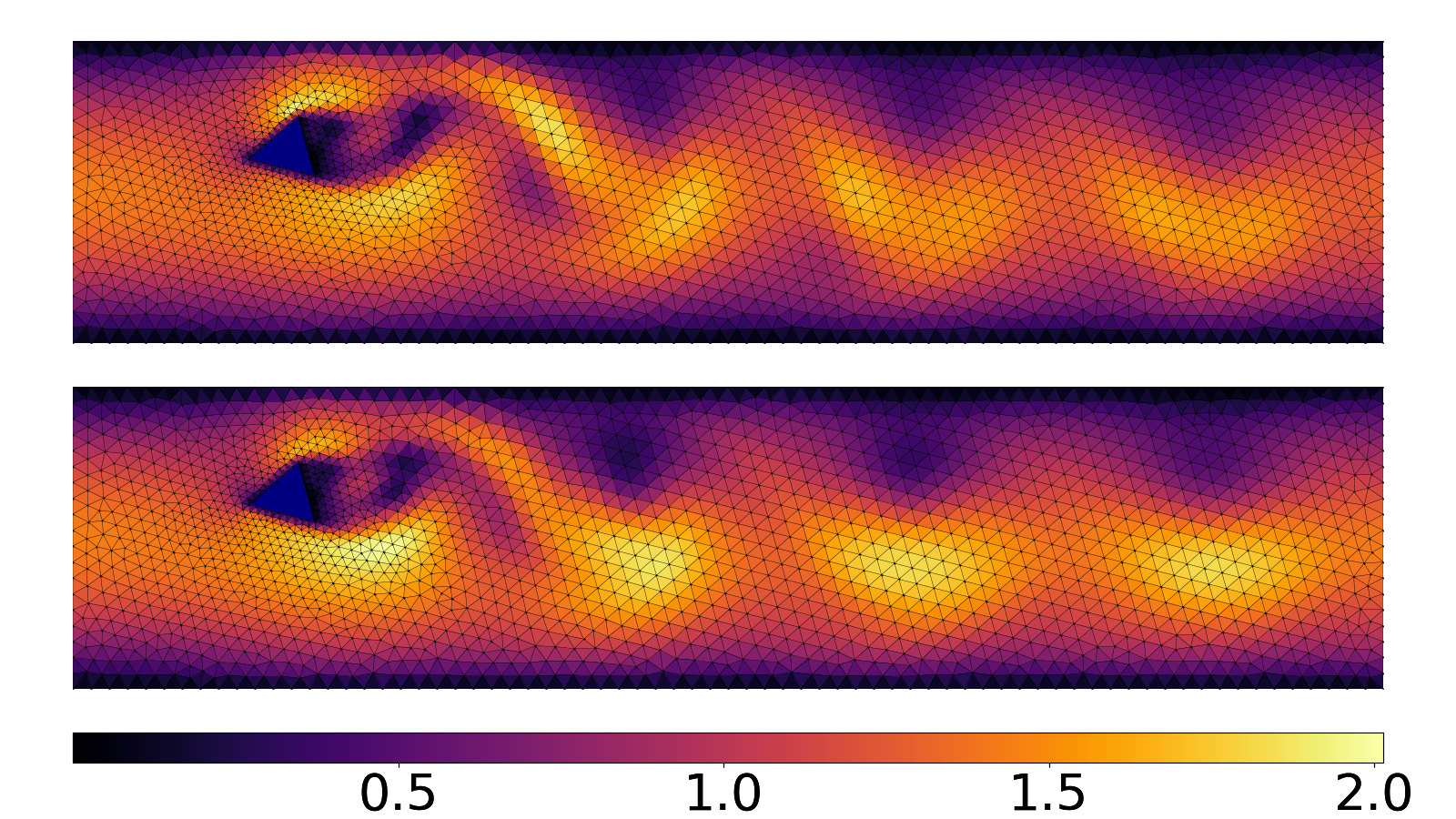}}
\hfill
\subfloat[][Dataset: \textbf{mixed\_all}, RMSE: $0.558$. Vortex pattern \textbf{falsely} predicted.]
{\includegraphics[width=0.47\textwidth]{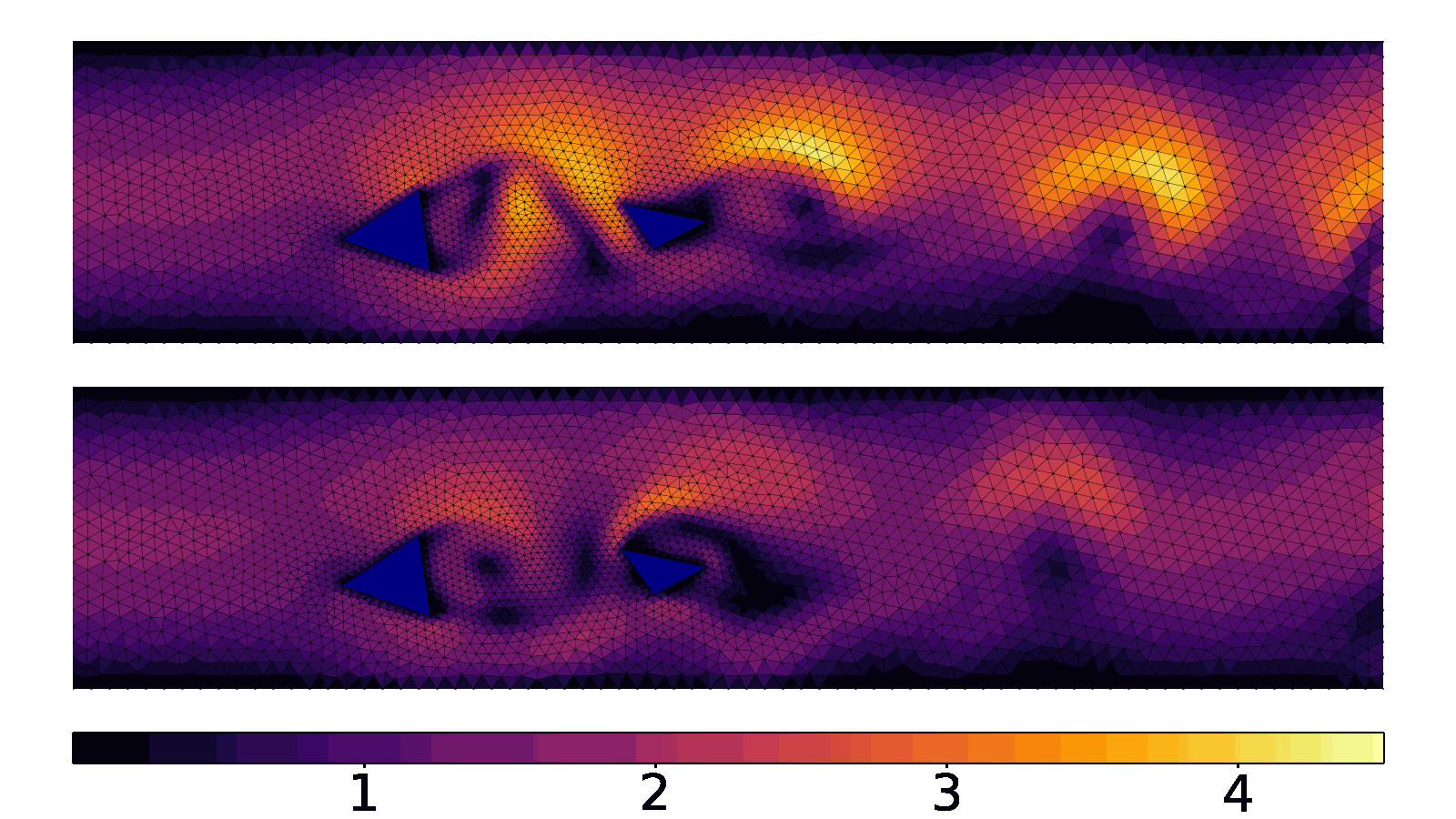}}
\hfill
\subfloat[][Dataset: \textbf{2cylinders}, RMSE: $0.389$. Vortex pattern \textbf{correctly} predicted.]
{\includegraphics[width=0.47\textwidth]{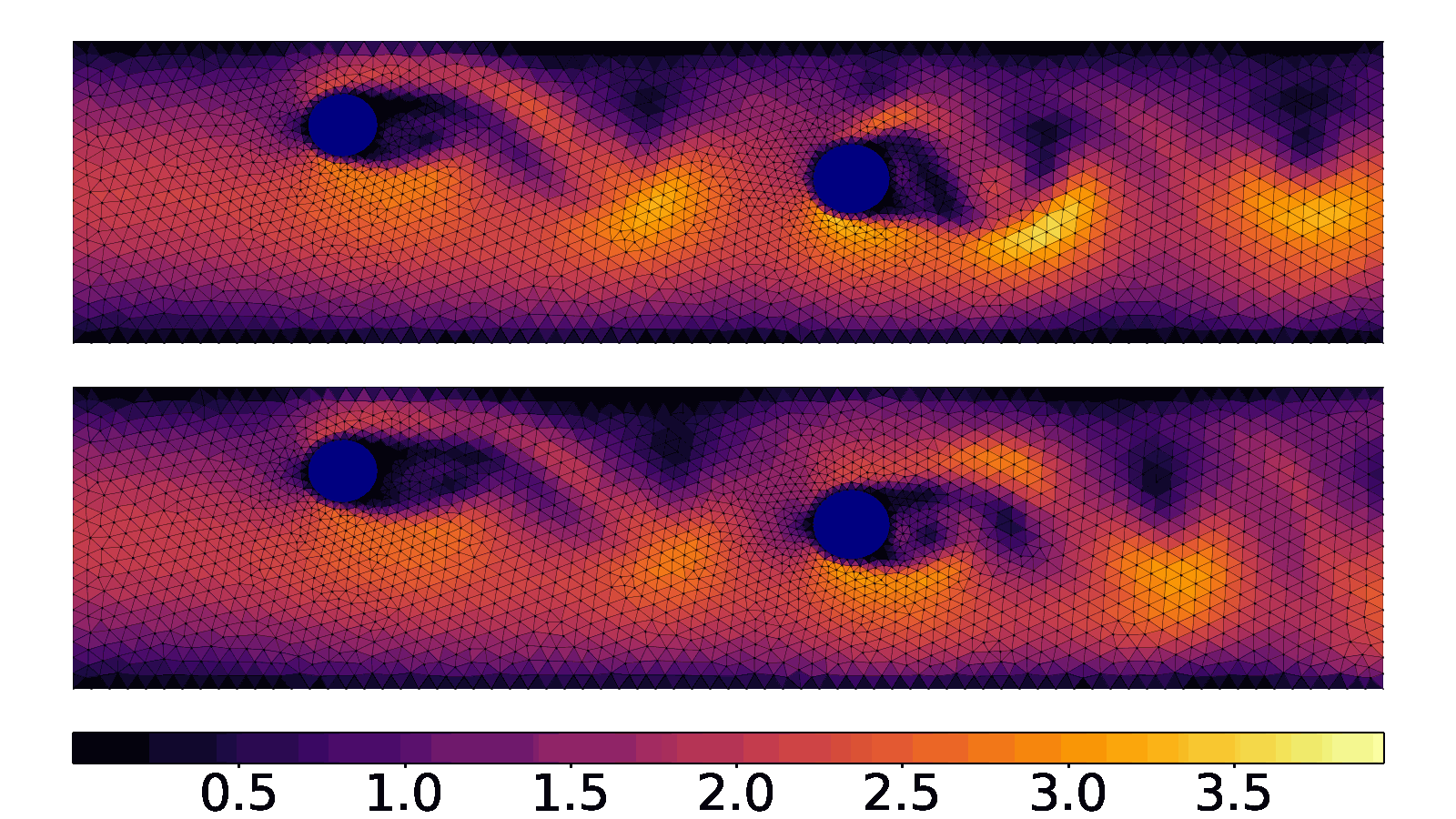}}
\hfill
\subfloat[][Dataset: \textbf{2cylinders}, RMSE: $0.318$. Vortex pattern \textbf{falsely} predicted.]
{\includegraphics[width=0.47\textwidth]{images/sims/standard_2cyl/sim10.png}}
\hfill
\centering
\subfloat[][Dataset: \textbf{cylinder\_stretch}, RMSE: $0.055$. Vortex pattern \textbf{correctly} predicted.]
{\includegraphics[width=0.47\textwidth]{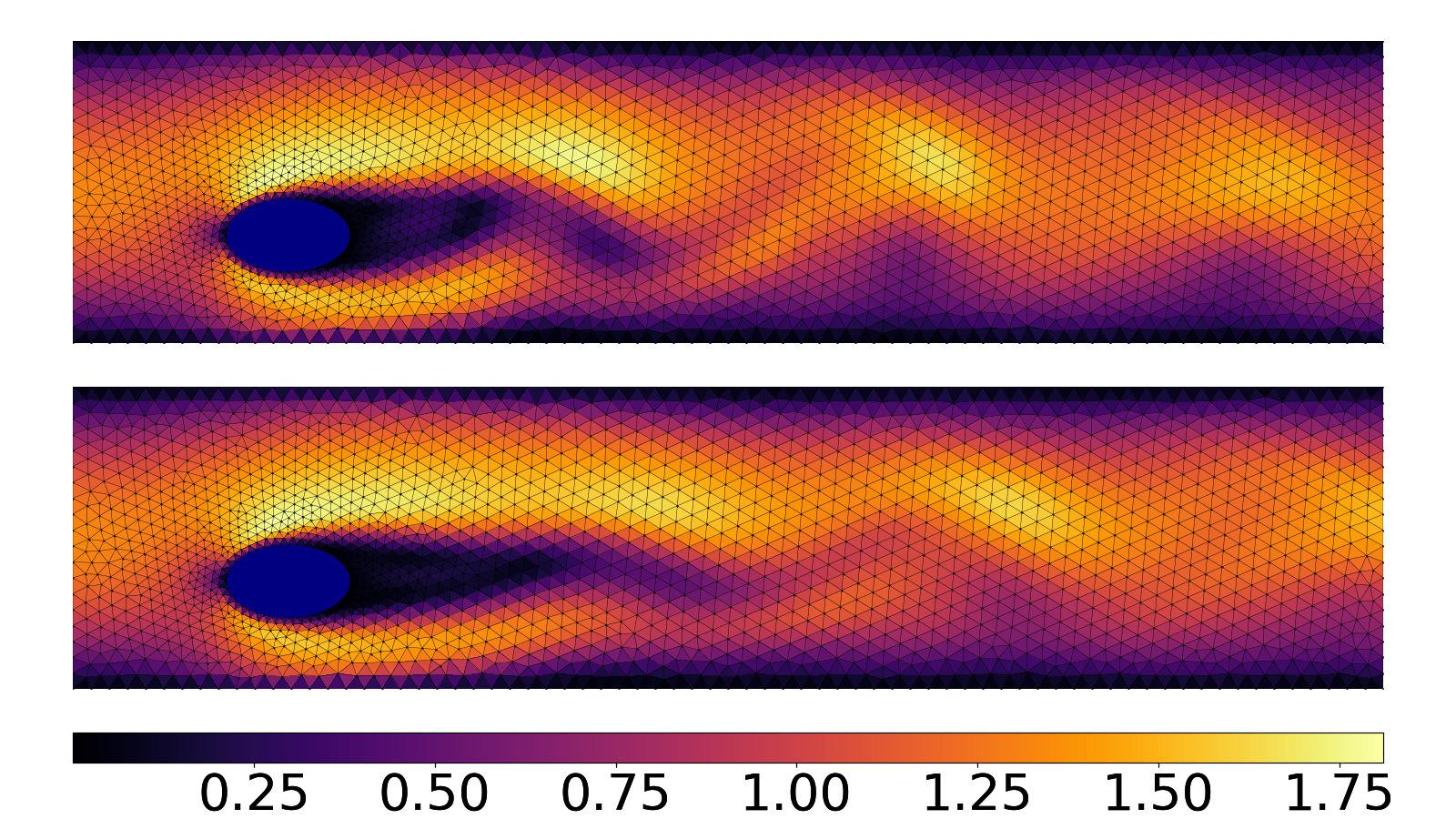}}

\caption{Plot of the Euclidean norm of the velocity field at the final time step for three FEM simulations (bottom) and their corresponding MGN predictions (top) of six meshes coming from the dataset mentioned in the corresponding subcaption. Each subcaption also contains the all-steps RMSE for the corresponding simulation rounded to three decimal places. All predictions stem from the same MGN that was trained on \textbf{standard\_cylinder}.
}
\label{fig:diff_set_sims}
\end{figure}

Though the RMSE values in all metrics do not differ much in relative terms between different training-evaluation dataset pairs, we can observe discrepancies when looking at the simulations. Fig. \ref{fig:diff_set_sims} shows predictions of a model trained on \textbf{standard\_cylinder} and evaluated on a selection of simulations from different datasets. This figure shows that if vortex shedding is predicted, it is essentially a coin flip whether the correct vortex pattern is predicted on a previously unseen shape. This is not reflected in the RMSE as an out-of-sync but characteristically correct vortex prediction can have the same RMSE as a straight-up wrong pattern.

\subsubsection{Intermediate conclusions}
Judging purely from the reported RMSE values, one could argue that MGNs generalize well for all training-evaluation dataset pairs. All training-evaluation dataset pairs' velocity RMSE is almost always an order of magnitude below what a freshly initialized MGN would produce in all metrics. For most datasets (unsurprisingly) the model that was evaluated and trained on the same dataset performs best. However, the other models which were trained on different datasets perform only slightly worse in relative terms. 

In general, MGNs are strong at predicting flow on a coarse level, therefore they almost always produce a physically reasonable prediction even for unseen shapes. However, their main weakness is correctly predicting the vortex patterns which is arguably the most difficult and interesting part of the prediction. This problem is more likely to occur when the mesh contains a shape that was never seen in training. However, we do note that though not always, MGNs can oftentimes correctly predict a vortex pattern for a shape that was not seen in training.

Given that our experiment setup is not optimal since one could use a larger MGN, finer meshes, more precisely generated datasets, one could achieve a significantly higher accuracy which is likely to translate to the generalization tasks. Therefore, one has to take into consideration that these experiments do not show the full potential of MGNs.

\subsection{Generalization experiments: Runtimes}
\label{sec:exp_runtimes}
In this section, we compare the runtime of an MGN to that of the IPCS on \textbf{standard\_cylinder} using the implementation and hyperparameters we described in the experiment setup. For the MGN we differentiate between using a GPU and only using a CPU. The aim is to find the number of simulations one wants to generate at which it is computationally more efficient to use an MGN over the IPCS. All runtimes reported in the following are extrapolations.

Firstly, we consider the case where we only use a CPU. In this case, we consider a system with six Intel(R) Core(TM) i5-9600K @ 3.70GHz CPUs with 32 GB of RAM. Using this system, we need roughly $0.3125$h to generate a single simulation using the IPCS which results in circa $125$h to generate the entire $\textbf{standard\_cylinder}$ train dataset. We then need roughly $1138$h to train the MGN and then approximately $0.048$h to generate a new simulation using the MGN. At inference time, this is an approximately \bm{$6.5 $} \textbf{times speedup}. 

Next, we consider the case where we have access to a GPU for training and inference of the MGN. Keeping everything as above and using the Alienware m17 R5 AMD laptop with the GeForce RTX 3070 Ti as the GPU with 8 GB of RAM
only for the MGN training and inference,
we get a total MGN training time of roughly 58h and then approximately 0.0031h to generate a new simulation. At inference time, this is an approximately \bm{$100.8$} \textbf{times speedup}. 

\section{Conclusion}
\label{sec_conclusion}

In this work, we investigated MGNs by extending the generalization experiments of DeepMind \cite{MGN} by evaluating MGNs on meshes with previously unseen shapes. For this, we first created five datasets \textbf{standard\_cylinder}, \textbf{cylinder\_stretch}, \textbf{cylinder\_tri\_quad}, \textbf{2cylinders}, and \textbf{mixed\_all}, which contain simulations of fluid flow around one or two objects of different shapes. We showed that MGNs in general can produce physically reasonable predictions for meshes with unseen shapes. Additionally, our MGNs managed to correctly predict vortex patterns for many unseen shapes. However, when a shape has not yet been seen in training the MGN is equally likely to falsely predict the vortex pattern for inflow peaks at which vortex shedding occurs. Nevertheless, for many engineering appliactions it is sufficient to have a coarse approximation of the flow field and the inference of the MGN is in our simulations up to $100$ times faster than the classical numerical solver.

In the future, one could consider differently shaped domains as opposed to the rectangular one chosen here and other fluid benchmark problems like lid-driven cavity or backward facing step. Additionally, one could also consider adding system parameters like the viscosity to the MGN and consider varying it during training and see how well the MGN generalizes in this case.

\section{Declarations}

\subsection{Availability of data and material}
The code for this article is available at \url{https://github.com/codebro634/modulus} \cite{github_implementation}. All other data will be made 
available on reasonable request.

\subsection{Competing interests}
The authors declare that they have no known competing financial interests or personal relationships that could have appeared to influence the work reported in this paper.

\subsection{Funding and acknowledgements}
The third and fourth authors acknowledge the funding of the German Research Foundation (DFG)
within the framework of the International Research Training Group on
Computational Mechanics Techniques in High Dimensions GRK 2657 under Grant
Number 433082294. The second author acknowledges support
by an ETH Zurich Postdoctoral Fellowship.


\subsection{Credit Authorship Contribution Statement}
{\color{black}
\makebox[2.75cm][l]{\textbf{R. Schmöcker}:} Conceptualization, Methodology, Software, Validation, Formal analysis, \\ 
\makebox[2.75cm][l]{} Investigation, Writing – original draft, Visualization. \\
\makebox[2.75cm][l]{\textbf{A. Henkes}:} Validation, Formal analysis, Investigation, Writing – original draft, \\ 
\makebox[2.75cm][l]{} Writing – review \& editing, Advisory. \\
\makebox[2.75cm][l]{\textbf{J. Roth}:} Conceptualization, Methodology, Validation, Formal analysis, Investigation,
\\ 
\makebox[2.75cm][l]{} Writing – original draft, Writing – review \& editing, Visualization, Supervision. \\
\makebox[2.75cm][l]{\textbf{T. Wick}:} Conceptualization, Formal analysis, Resources, Writing - review \& editing,
\\ 
\makebox[2.75cm][l]{} Supervision, Funding acquisition. \\
}


\bibliographystyle{abbrv}
\bibliography{lit.bib}

\begin{thebibliography}{10}

\bibitem{aggarwal2018neural}
C.~C. Aggarwal.
\newblock {\em Neural Networks and Deep Learning: A Textbook}.
\newblock Springer International Publishing, 2018.

\bibitem{Allen2022}
K.~R. Allen, T.~Lopez-Guevara, K.~L. Stachenfeld, A.~Sanchez-Gonzalez, P.~W.
  Battaglia, J.~B. Hamrick, and T.~Pfaff.
\newblock Inverse {D}esign for {F}luid-{S}tructure {I}nteractions using {G}raph
  {N}etwork {S}imulators.
\newblock In {\em Neural Information Processing Systems}, 2022.

\bibitem{fenicsproject}
M.~S. Alnaes, J.~Blechta, J.~Hake, A.~Johansson, B.~Kehlet, A.~Logg,
  C.~Richardson, J.~Ring, M.~E. Rognes, and G.~N. Wells.
\newblock {The FEniCS Project Version 1.5}.
\newblock {\em Archive of Numerical Software}, 3, 2015.

\bibitem{Amsallem_Zahr_Farhat_2012}
D.~Amsallem, M.~J. Zahr, and C.~Farhat.
\newblock Nonlinear model order reduction based on local reduced-order bases.
\newblock {\em Int. J. Numer. Methods Eng.}, 92(10):891--916, 2012.

\bibitem{Anselmann2023}
M.~Anselmann and M.~Bause.
\newblock A {G}eometric {M}ultigrid {M}ethod for {S}pace-{T}ime {F}inite
  {E}lement {D}iscretizations of the {N}avier–{S}tokes {E}quations and its
  {A}pplication to 3{D} {F}low {S}imulation.
\newblock {\em ACM Trans. Math. Softw.}, 49(1), mar 2023.

\bibitem{ExaDG}
D.~Arndt, N.~Fehn, G.~Kanschat, K.~Kormann, M.~Kronbichler, P.~Munch, W.~A.
  Wall, and J.~Witte.
\newblock Exa{D}{G}: {H}igh-{O}rder {D}iscontinuous {G}alerkin for the
  {E}xa-{S}cale.
\newblock In H.-J. Bungartz, S.~Reiz, B.~Uekermann, P.~Neumann, and W.~E.
  Nagel, editors, {\em Software for Exascale Computing - SPPEXA 2016-2019},
  pages 189--224, Cham, 2020. Springer International Publishing.

\bibitem{Astrid2008MissingPE}
P.~Astrid, S.~Weiland, K.~Willcox, and T.~Backx.
\newblock {M}issing {P}oint {E}stimation in {M}odels {D}escribed by {P}roper
  {O}rthogonal {D}ecomposition.
\newblock {\em IEEE Trans. Autom. Control}, 53:2237--2251, 2008.

\bibitem{ba2016layer}
J.~L. Ba, J.~R. Kiros, and G.~E. Hinton.
\newblock {Layer Normalization}, 2016.
\newblock \url{https://doi.org/10.48550/arXiv.1607.06450}.

\bibitem{Barrault2004AnI}
M.~Barrault, Y.~Maday, N.~Nguyen, and A.~Patera.
\newblock An ‘empirical interpolation’ method: application to efficient
  reduced-basis discretization of partial differential equations.
\newblock {\em C.R. Math.}, 339:667--672, 2004.

\bibitem{BeRa2001}
R.~Becker and R.~Rannacher.
\newblock An optimal control approach to a posteriori error estimation in
  finite element methods.
\newblock {\em Acta Numerica}, 10:1–102, 2001.

\bibitem{BeRa12}
M.~Besier and R.~Rannacher.
\newblock Goal-oriented space-time adaptivity in the finite element {G}alerkin
  method for the computation of nonstationary incompressible flow.
\newblock {\em Int. J. Num. Meth. Fluids}, 70:1139--1166, 2012.

\bibitem{bishop2006pattern}
C.~M. Bishop.
\newblock {\em Pattern recognition and machine learning}.
\newblock Springer, 2006.

\bibitem{bonnet2022}
F.~Bonnet, J.~A. Mazari, T.~Munzer, P.~Yser, and P.~Gallinari.
\newblock An extensible {B}enchmarking {G}raph-{M}esh dataset for studying
  {S}teady-{S}tate {I}ncompressible {N}avier-{S}tokes {E}quations.
\newblock In {\em ICLR 2022 Workshop on Geometrical and Topological
  Representation Learning}, 2022.

\bibitem{Braack2006}
M.~Braack and T.~Richter.
\newblock Solutions of 3{D} {N}avier–{S}tokes benchmark problems with
  adaptive finite elements.
\newblock {\em Comput. Fluids}, 35(4):372--392, 2006.

\bibitem{bricken2023monosemanticity}
T.~Bricken, A.~Templeton, J.~Batson, B.~Chen, A.~Jermyn, T.~Conerly, N.~Turner,
  C.~Anil, C.~Denison, A.~Askell, R.~Lasenby, Y.~Wu, S.~Kravec, N.~Schiefer,
  T.~Maxwell, N.~Joseph, Z.~Hatfield-Dodds, A.~Tamkin, K.~Nguyen, B.~McLean,
  J.~E. Burke, T.~Hume, S.~Carter, T.~Henighan, and C.~Olah.
\newblock Towards {M}onosemanticity: {D}ecomposing {L}anguage {M}odels {W}ith
  {D}ictionary {L}earning.
\newblock {\em Transformer Circuits Thread}, 2023.
\newblock
  \url{https://transformer-circuits.pub/2023/monosemantic-features/index.html}.

\bibitem{Bronstein2021Book}
M.~M. Bronstein, J.~Bruna, T.~Cohen, and P.~Veličković.
\newblock Geometric {D}eep {L}earning: {G}rids, {G}roups, {G}raphs,
  {G}eodesics, and {G}auges, 2021.
\newblock \url{https://doi.org/10.48550/arXiv.2104.13478}.

\bibitem{Bronstein2017}
M.~M. Bronstein, J.~Bruna, Y.~LeCun, A.~Szlam, and P.~Vandergheynst.
\newblock Geometric {D}eep {L}earning: {G}oing beyond {E}uclidean data.
\newblock {\em IEEE Signal Processing Magazine}, 34(4):18--42, 2017.

\bibitem{Cai2021}
S.~Cai, Z.~Mao, Z.~Wang, M.~Yin, and G.~E. Karniadakis.
\newblock Physics-informed neural networks ({P}{I}{N}{N}s) for fluid mechanics:
  a review.
\newblock {\em Acta Mech. Sin.}, 37(12):1727--1738, Dec 2021.

\bibitem{Chaturantabut2010NonlinearMR}
S.~Chaturantabut and D.~Sorensen.
\newblock {N}onlinear {M}odel {R}eduction via {D}iscrete {E}mpirical
  {I}nterpolation.
\newblock {\em SIAM J. Sci. Comput.}, 32:2737--2764, 2010.

\bibitem{Chaturantabut2012ASS}
S.~Chaturantabut and D.~Sorensen.
\newblock {A} {S}tate {S}pace {E}rror {E}stimate for {P}{O}{D}-{D}{E}{I}{M}
  {N}onlinear {M}odel {R}eduction.
\newblock {\em SIAM J. Numer. Anal.}, 50:46--63, 2012.

\bibitem{Chen2021}
J.~Chen, E.~Hachem, and J.~Viquerat.
\newblock {Graph neural networks for laminar flow prediction around random
  two-dimensional shapes}.
\newblock {\em Phys. Fluids}, 33(12):123607, 12 2021.

\bibitem{chollet2018deep}
F.~Chollet.
\newblock {\em Deep {L}earning with {P}ython}, volume 361.
\newblock Manning New York, 2018.

\bibitem{chorinsmethod}
A.~J. Chorin.
\newblock {Numerical Solution of the Navier-Stokes Equations}.
\newblock {\em Math. Comput.}, 22(104):745--762, 1968.

\bibitem{Q-DEIM}
Z.~Drmač and S.~Gugercin.
\newblock {A} {N}ew {S}election {O}perator for the {D}iscrete {E}mpirical
  {I}nterpolation {M}ethod---{I}mproved {A} {P}riori {E}rror {B}ound and
  {E}xtensions.
\newblock {\em SIAM J. Sci. Comput.}, 38(2):A631--A648, 2016.

\bibitem{ElSiWa14}
H.~Elman, D.~Silvester, and A.~Wathen.
\newblock {\em Finite Elements and Fast Iterative Solvers}.
\newblock Oxford University Press, 2014.

\bibitem{Erhan2009}
D.~Erhan, Y.~Bengio, A.~Courville, and P.~Vincent.
\newblock Visualizing {H}igher-{L}ayer {F}eatures of a {D}eep {N}etwork.
\newblock {\em Technical Report, Univeristé de Montréal}, 01 2009.

\bibitem{MSMGN}
M.~Fortunato, T.~Pfaff, P.~Wirnsberger, A.~Pritzel, and P.~Battaglia.
\newblock {MultiScale MeshGraphNets}.
\newblock In {\em ICML 2022 2nd AI for Science Workshop}, 2022.

\bibitem{galdi2011}
G.~Galdi.
\newblock {\em An Introduction to the Mathematical Theory of the Navier-Stokes
  Equations: Steady-State Problems}.
\newblock Springer Monographs in Mathematics. Springer New York, 2011.

\bibitem{geron2019hands}
A.~G{\'e}ron.
\newblock {\em Hands-on machine learning with Scikit-Learn, Keras, and
  TensorFlow: Concepts, tools, and techniques to build intelligent systems}.
\newblock O'Reilly Media, 2019.

\bibitem{GiRa1986}
V.~Girault and P.-A. Raviart.
\newblock {\em Finite Element method for the {N}avier-{S}tokes equations}.
\newblock Number 5 in Computer Series in Computational Mathematics.
  Springer-Verlag, 1986.

\bibitem{Glowinski2003}
R.~Glowinski.
\newblock Finite element methods for incompressible viscous flow.
\newblock In {\em Numerical Methods for Fluids (Part 3)}, volume~9 of {\em
  Handbook of Numerical Analysis}, pages 3--1176. Elsevier, 2003.

\bibitem{goodfellow2016deep}
I.~Goodfellow, Y.~Bengio, A.~Courville, and Y.~Bengio.
\newblock {\em Deep learning}, volume~1.
\newblock MIT press Cambridge, 2016.

\bibitem{Graessle2019}
C.~Gr{\"a}{\ss}le, M.~Hinze, J.~Lang, and S.~Ullmann.
\newblock {P}{O}{D} model order reduction with space-adapted snapshots for
  incompressible flows.
\newblock {\em Adv. Comput. Math.}, 45:2401--2428, 2019.

\bibitem{Grimm2023}
V.~Grimm, A.~Heinlein, and A.~Klawonn.
\newblock Learning the solution operator of two-dimensional incompressible
  {N}avier-{S}tokes equations using physics-aware convolutional neural
  networks, 2023.
\newblock \url{https://doi.org/10.48550/arXiv.2308.02137}.

\bibitem{hauser2018principles}
M.~B. Hauser.
\newblock Principles of {R}iemannian geometry in neural networks.
\newblock {\em PhD thesis}, 2018.

\bibitem{NVIDIA_SimNet}
O.~Hennigh, S.~Narasimhan, M.~A. Nabian, A.~Subramaniam, K.~Tangsali,
  M.~Rietmann, J.~del Aguila~Ferrandis, W.~Byeon, Z.~Fang, and S.~Choudhry.
\newblock {NVIDIA} {SimNet}: an {AI}-accelerated multi-physics simulation
  framework, 2020.
\newblock \url{https://doi.org/10.48550/arXiv.2012.07938}.

\bibitem{Ubbiali2018}
J.~Hesthaven and S.~Ubbiali.
\newblock Non-intrusive reduced order modeling of nonlinear problems using
  neural networks.
\newblock {\em J. Comput. Phys.}, 363:55--78, 2018.

\bibitem{hornik1989multilayer}
K.~Hornik, M.~Stinchcombe, and H.~White.
\newblock Multilayer feedforward networks are universal approximators.
\newblock {\em Neural networks}, 2(5):359--366, 1989.

\bibitem{John1999}
V.~John.
\newblock A comparison of parallel solvers for the incompressible
  {N}avier--{S}tokes equations.
\newblock {\em Comput. Visualization Sci.}, 1(4):193--200, Jul 1999.

\bibitem{John2016}
V.~John.
\newblock {\em Finite Element Methods for Incompressible Flow Problems}.
\newblock Springer, 2016.

\bibitem{kenningtondifferential}
A.~U. Kennington.
\newblock Differential geometry reconstructed: a unified systematic framework.
\newblock \url{http://www.topology.org/tex/conc/dg.html}, 2024.

\bibitem{Kochkov2021}
D.~Kochkov, J.~A. Smith, A.~Alieva, Q.~Wang, M.~P. Brenner, and S.~Hoyer.
\newblock Machine learning–accelerated computational fluid dynamics.
\newblock {\em PNAS}, 118(21):e2101784118, 2021.

\bibitem{AlexNet2012}
A.~Krizhevsky, I.~Sutskever, and G.~E. Hinton.
\newblock Imagenet classification with deep convolutional neural networks.
\newblock In F.~Pereira, C.~Burges, L.~Bottou, and K.~Weinberger, editors, {\em
  Advances in Neural Information Processing Systems}, volume~25. Curran
  Associates, Inc., 2012.

\bibitem{kutz_2017}
J.~N. Kutz.
\newblock Deep learning in fluid dynamics.
\newblock {\em J. Fluid Mech.}, 814:1–4, 2017.

\bibitem{graphcast}
R.~Lam, A.~Sanchez-Gonzalez, M.~Willson, P.~Wirnsberger, M.~Fortunato, F.~Alet,
  S.~Ravuri, T.~Ewalds, Z.~Eaton-Rosen, W.~Hu, A.~Merose, S.~Hoyer, G.~Holland,
  O.~Vinyals, J.~Stott, A.~Pritzel, S.~Mohamed, and P.~Battaglia.
\newblock {Learning skillful medium-range global weather forecasting}.
\newblock {\em Science}, 382(6677):1416--1421, 2023.

\bibitem{fenicstutorial}
H.~P. Langtangen and A.~Logg.
\newblock {\em {Solving PDEs in Python}}.
\newblock Simula SpringerBriefs on Computing. Springer Cham, 1 edition, 2017.

\bibitem{GINO2023}
Z.~Li, N.~B. Kovachki, C.~Choy, B.~Li, J.~Kossaifi, S.~P. Otta, M.~A. Nabian,
  M.~Stadler, C.~Hundt, K.~Azizzadenesheli, and A.~Anandkumar.
\newblock Geometry-{I}nformed {N}eural {O}perator for {L}arge-{S}cale 3{D}
  {PDE}s.
\newblock In {\em Thirty-seventh Conference on Neural Information Processing
  Systems}, 2023.

\bibitem{Margenberg2022}
N.~Margenberg, D.~Hartmann, C.~Lessig, and T.~Richter.
\newblock A neural network multigrid solver for the {N}avier-{S}tokes
  equations.
\newblock {\em J. Comput. Phys.}, 460:110983, 2022.

\bibitem{MiRaHaTe91}
S.~Mittal, A.~Ratner, D.~Hastreiter, and T.~Tezduyar.
\newblock Space-time finite element computation of incompressible flows with
  emphasis on flow involving oscillating cylinders.
\newblock {\em Internat. Video J. Engrg. Res.}, 1:83--86, 1991.

\bibitem{MouMaDa16}
F.~Moukalled, L.~Mangani, and M.~Darwish.
\newblock {\em The Finite Volume Method in Computational Fluid Dynamics}.
\newblock Springer, 2016.

\bibitem{Nguyen_patera_peraire_2008}
N.~Nguyen, A.~Patera, and J.~Peraire.
\newblock A ‘best points’ interpolation method for efficient approximation
  of parametrized function.
\newblock {\em Int. J. Numer. Methods Eng.}, 73:521 -- 543, 01 2008.

\bibitem{novak2018sensitivity}
R.~Novak, Y.~Bahri, D.~A. Abolafia, J.~Pennington, and J.~Sohl-Dickstein.
\newblock Sensitivity and {G}eneralization in {N}eural {N}etworks: an
  {E}mpirical {S}tudy.
\newblock In {\em International Conference on Learning Representations}, 2018.

\bibitem{NVIDIA_Modulus}
{\relax NVIDIA Modulus Team}.
\newblock {NVIDIA} {Modulus}.
\newblock \url{https://github.com/NVIDIA/modulus/tree/main}.

\bibitem{MGN}
T.~Pfaff, M.~Fortunato, A.~Sanchez-Gonzalez, and P.~Battaglia.
\newblock Learning {M}esh-{B}ased {S}imulation with {G}raph {N}etworks.
\newblock In {\em International Conference on Learning Representations}, 2021.

\bibitem{Rannacher2000}
R.~Rannacher.
\newblock {\em Finite Element Methods for the Incompressible Navier-Stokes
  Equations}, pages 191--293.
\newblock Birkh{\"a}user Basel, Basel, 2000.

\bibitem{RoThiKoeWi2024}
J.~Roth, J.~P. Thiele, U.~Köcher, and T.~Wick.
\newblock Tensor-{P}roduct {S}pace-{T}ime {G}oal-{O}riented {E}rror {C}ontrol
  and {A}daptivity {W}ith {P}artition-of-{U}nity {D}ual-{W}eighted {R}esiduals
  for {N}onstationary {F}low {P}roblems.
\newblock {\em Comput. Methods Appl. Math.}, 24(1):185--214, 2024.

\bibitem{Rozza2007Error}
G.~Rozza, D.~B.~P. Huynh, and A.~T. Patera.
\newblock Reduced basis approximation and a posteriori error estimation for
  affinely parametrized elliptic coercive partial differential equations.
\newblock {\em Arch. Comput. Methods Eng.}, 15(3):1--47, Sep 2007.

\bibitem{Rozza2022CFD}
G.~Rozza, G.~Stabile, and F.~Ballarin.
\newblock {\em Advanced Reduced Order Methods and Applications in Computational
  Fluid Dynamics}.
\newblock Society for Industrial and Applied Mathematics, Philadelphia, PA,
  2022.

\bibitem{MGN_pre_paper}
A.~Sanchez-Gonzalez, J.~Godwin, T.~Pfaff, R.~Ying, J.~Leskovec, and
  P.~Battaglia.
\newblock {Learning to Simulate Complex Physics with Graph Networks}.
\newblock In H.~D. III and A.~Singh, editors, {\em Proceedings of the 37th
  International Conference on Machine Learning}, volume 119 of {\em Proceedings
  of Machine Learning Research}, pages 8459--8468. PMLR, 13--18 Jul 2020.

\bibitem{lengeling2021}
B.~Sanchez-Lengeling, E.~Reif, A.~Pearce, and A.~B. Wiltschko.
\newblock A {G}entle {I}ntroduction to {G}raph {N}eural {N}etworks.
\newblock {\em Distill}, 2021.
\newblock \url{https://distill.pub/2021/gnn-intro}.

\bibitem{pygmsh}
N.~Schlömer.
\newblock {pygmsh: A Python frontend for Gmsh}.
\newblock \url{https://github.com/nschloe/pygmsh} [Last accessed 10.05.2024],
  2018.

\bibitem{dfg_benchmark}
M.~Schäfer, S.~Turek, F.~Durst, E.~Krause, and R.~Rannacher.
\newblock {Benchmark Computations of Laminar Flow Around a Cylinder}.
\newblock {\em Vieweg+Teubner Verlag}, pages 547--566, 1996.

\bibitem{Simonyan14a}
K.~Simonyan, A.~Vedaldi, and A.~Zisserman.
\newblock Deep {I}nside {C}onvolutional {N}etworks: {V}isualising {I}mage
  {C}lassification {M}odels and {S}aliency {M}aps.
\newblock In {\em Workshop at International Conference on Learning
  Representations}, 2014.

\bibitem{Tang2023TowardsUG}
H.~Tang and Y.~Liu.
\newblock Towards {U}nderstanding {G}eneralization of {G}raph {N}eural
  {N}etworks.
\newblock In {\em International Conference on Machine Learning}, 2023.

\bibitem{github_implementation}
K.~Tangsali, M.~A. Nabian, N.~Geneva, R.~Schmöcker, A.~Subramaniam, D.~Foster,
  O.~Hennigh, D.~Pruitt, M.~Stadler, R.~Cherukuri, M.~Koch, S.~Wang, N.~D.
  Brenowitz, A.~Kamenev, B.~M.~L. Borgne, J.~Yang, L.~Pegolotti, M.~Mardani,
  S.~Jones, S.~Roy, and T.~Kurth.
\newblock codebro634/modulus: v1.0.0.
\newblock \url{https://doi.org/10.5281/zenodo.11504394}, June 2024.

\bibitem{temam2001}
R.~Temam.
\newblock {\em Navier-Stokes Equations: Theory and Numerical Analysis}.
\newblock AMS/Chelsea publication. AMS Chelsea Pub., 2001.

\bibitem{TeBeMiJo92}
T.~E. Tezduyar, M.~Behr, S.~Mittal, and A.~A. Johnson.
\newblock {\em Computation of unsteady incompressible flows with the Finite
  Element Methods Space-Time Formulations, Iterative Strategies and Massively
  Parallel Implementations}, volume 143 of {\em New Methods in Transient
  Analysis, PVP-Vol. 246, AMD-Vol. 143}, pages 7--24.
\newblock ASME, New York, 1992.

\bibitem{turek1999}
S.~Turek.
\newblock {\em Efficient Solvers for Incompressible Flow Problems: An
  Algorithmic and Computational Approach}.
\newblock Lecture Notes in Computational Science and Engineering.
  Springer-Verlag, 1999.

\bibitem{wu2024Transolver}
H.~Wu, H.~Luo, H.~Wang, J.~Wang, and M.~Long.
\newblock Transolver: {A} {F}ast {T}ransformer {S}olver for {PDE}s on {G}eneral
  {G}eometries.
\newblock In {\em International Conference on Machine Learning}, 2024.

\bibitem{Zeiler2014}
M.~D. Zeiler and R.~Fergus.
\newblock Visualizing and {U}nderstanding {C}onvolutional {N}etworks.
\newblock In D.~Fleet, T.~Pajdla, B.~Schiele, and T.~Tuytelaars, editors, {\em
  Computer Vision -- ECCV 2014}, pages 818--833, Cham, 2014. Springer
  International Publishing.

\end{thebibliography}

\end{document}